\documentclass[sigconf]{acmart}
\usepackage{balance}

\copyrightyear{2025} 
\acmYear{2025} 
\setcopyright{acmlicensed}
\acmConference[WWW '25]{Proceedings of the ACM
 Web Conference 2025}{April 28-May 2, 2025}{Sydney, NSW, Australia}
\acmBooktitle{Proceedings of the ACM Web Conference 2025 (WWW '25), April
 28-May 2, 2025, Sydney, NSW, Australia}
\acmDOI{10.1145/3696410.3714749}
\acmISBN{979-8-4007-1274-6/25/04}
\usepackage[utf8]{inputenc} 
\usepackage[T1]{fontenc}    
\usepackage{hyperref}       
\usepackage{url}            
\usepackage{booktabs}       
\usepackage{amsfonts}       
\usepackage{nicefrac}       
\usepackage{microtype}      
\usepackage{xcolor}         

\usepackage{graphicx}
\usepackage{multirow} 

\usepackage{algorithm}
\usepackage{algorithmic}
\usepackage{subfigure}
\usepackage{mathtools}

\usepackage{amssymb,amsmath,amsthm}

\newtheorem{condition}{Condition}
\usepackage{subcaption}
\begin{document}

\title{Leveraging Invariant Principle for Heterophilic Graph Structure Distribution Shifts}

    \author{Jinluan Yang}
    \orcid{0009-0007-1844-7029}
    \affiliation{%
      \institution{Zhejiang University}
      \city{Hangzhou}
      \country{China}
    }
    \email{yangjinluan@zju.edu.cn}
    
    \author{Zhengyu Chen}
    \orcid{0000-0002-9863-556X}
    \affiliation{%
      \institution{Meituan}
      \city{Beijing}
      \country{China}
    }
    \email{chenzhengyu04@meituan.com}

    \author{Teng Xiao}
    \orcid{0000-0001-9180-1433}
    \affiliation{%
      \institution{The Pennsylvania State University}
      \city{Pennsylvania}
      \country{United States}
    }
     \email{tengxiao@psu.edu}

    \author{Yong Lin}
    \orcid{0000-0003-3525-6738}
    \affiliation{%
      \institution{HKUST}
      \city{Hongkong}
      \country{China}
    }
    \email{ylindf@connect.ust.hk}

    \author{Wenqiao Zhang}
    \orcid{0000-0002-5988-7609}
    \affiliation{%
    \institution{Zhejiang University}
    \city{Hangzhou}
    \country{China}
    }
    \email{wenqiaozhang@zju.edu.cn}

    \author{Kun Kuang}
    \orcid{0000-0001-7024-9790}
    \authornote{Corresponding author}
    \affiliation{%
      \institution{Zhejiang University}
      \city{Hangzhou}
      \country{China}
    }
    \email{kunkuang@zju.edu.cn}
\renewcommand{\shortauthors}{Jinluan Yang et al.}

\begin{abstract}
Heterophilic  Graph Neural Networks (HGNNs) have shown promising results for semi-supervised learning tasks on graphs. Notably, most real-world heterophilic graphs are composed of a mixture of nodes with different neighbor patterns, exhibiting local node-level homophilic and heterophilic structures. However, existing works are only devoted to designing better unified HGNN backbones for node classification tasks on heterophilic and homophilic graphs simultaneously, and their analyses of HGNN performance concerning nodes are only based on the determined data distribution without exploring the effect caused by the difference of structural pattern between training and testing nodes. How to learn invariant node representations on heterophilic graphs to handle this structure difference or distribution shifts remains unexplored. In this paper, we first discuss the limitations of previous graph-based invariant learning methods in addressing the heterophilic graph structure distribution shifts from the perspective of data augmentation. Then, we propose \textbf{HEI}, a framework capable of generating invariant node representations through incorporating \textbf{H}eterophily information, the node's estimated neighbor pattern, to infer latent \textbf{E}nvironments without augmentation, which are then used for \textbf{I}nvariant prediction. We provide detailed theoretical guarantees to clarify the reasonability of HEI. Extensive experiments on various benchmarks and backbones can also demonstrate the effectiveness and robustness of our method compared with existing state-of-the-art baselines. Our codes can be accessed through \href{https://github.com/Yangjinluan/HEI}{HEI}.
\end{abstract}


\maketitle
\section{Introduction}
Graph Neural Networks (GNNs) have emerged as prominent approaches for learning graph-structured representations through the aggregation mechanism that effectively combines feature information from neighboring nodes~\cite{zheng2022graph}. 
Previous GNNs primarily dealt with \emph{homophilic graphs}, where connected nodes tend to share similar features and labels~\cite{zhu2020beyond}. 
However, growing empirical evidence suggests that these GNNs' performance significantly deteriorates when dealing with \emph{heterophilic graphs}, where most nodes connect with others from different classes, even worse than the traditional neural networks~\cite{lim2021large}. An appealing way to address this issue is to tailor the heterophily property to GNNs, extending the range of neighborhood aggregation and reorganizing architecture~\cite{zheng2022graph}, known as the heterophilic GNNs (HGNNs).

\begin{figure*}
\begin{center}
\centering
\setlength{\abovecaptionskip}{0cm}   
\setlength{\belowcaptionskip}{0cm}   
\includegraphics[width=0.8\linewidth]{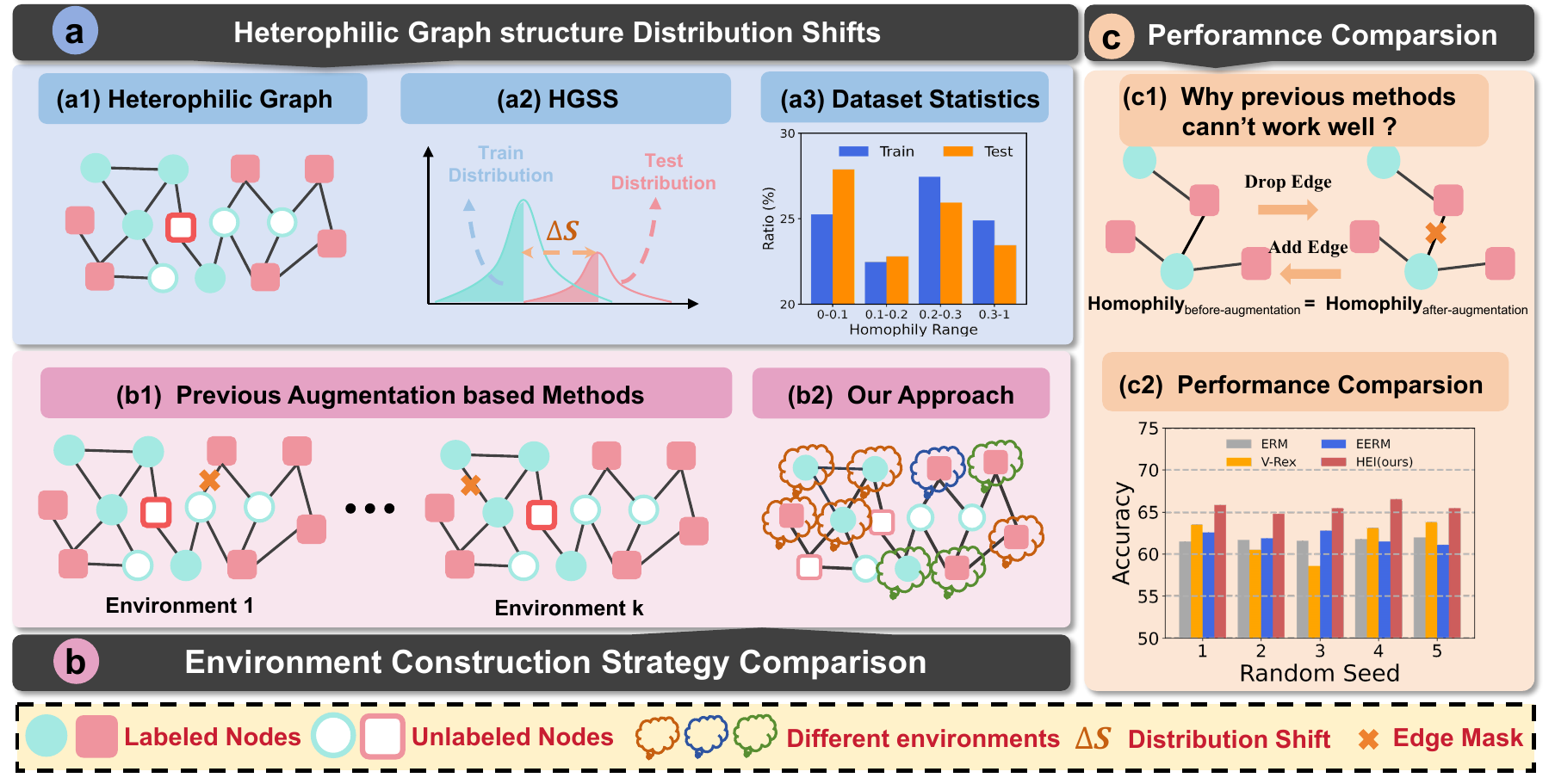}
\caption{(a) illustrates the heterophilic graph structure distribution shifts (HGSS), where the figure and histogram show the HGSS and neighbor pattern (measured by node homophily) difference between train and test nodes on the Squirrel dataset; (b) displays the comparison of different environment construction strategies between previous invariant learning works and ours from augmentation;
(c) shows that the environment construction of previous methods may be ineffective in addressing the HGSS due to the unchanged neighbor pattern distribution. The experimental results between traditional and graph-based invariant learning methods can support our analysis
and verify the superiority of our proposed HEI.} 

\Description{(a) illustrates the heterophilic graph structure distribution shifts (HGSS), where the figure and histogram show the HGSS and neighbor pattern (measured by node homophily) difference between train and test nodes on the Squirrel dataset; (b) displays the comparison of different environment construction strategies between previous works and ours, the key exists that we don't need to construct environments with augmentation; (c) shows that previous environment construction strategy may be ineffective in addressing our defined HGSS due to the unchanged neighbor pattern distribution. The experimental results between traditional and graph-based invariant learning methods can support our analysis.}
\label{intro}
\end{center}
\vspace{-0.2cm}
\end{figure*}


\emph{Heterophilic Graph Structure distribution Shift (HGSS): A novel data distribution shift perspective to reconsider existing HGNNs works.} Despite promising, most previous HGNNs assume the nodes share the determined data distribution \cite{lim2021large, li2022finding}, 
we argue that there is data distribution disparity among nodes with different neighbor patterns. As illustrated in Figure~\ref{intro} (a1), heterophilic graphs are composed of a mixture of nodes that exhibit local homophilic and heterophilic structures, \emph{i.e}, the nodes have different neighbor patterns~\cite{zheng2022graph}. The node's neighbor pattern can be measured by node homophily, representing homophily level by comparing the label between the node and its neighbors. Here, we identify their varying neighbor patterns between train and test nodes as the Heterophilic Graph Structure distribution Shift (Figure~\ref{intro} (a2)). This kind of shift was neglected by previous works but truly affected GNN's performance. As shown in Figure~\ref{intro} (a3), we visualize the HGSS between training and testing nodes on the Squirrel dataset. Compared with test nodes, the train nodes are more prone to be categorized into groups with high homophily, which may yield a test performance degradation. Notably, though some recent work \cite{mao2023demystifying} also discusses homophilic and heterophilic structural patterns, until now they haven't provided a clear technique solution for this problem. 
Compared with traditional HGNN works that focus on backbone designs, it's extremely urgent to seek solutions from a data distribution perspective to address the HGSS issue.

\emph{Existing graph-based invariant learning methods perform badly for HGSS due to the augmentation-based environment construction strategy.}
In the context of general distribution shifts, the technique of invariant learning~\cite{rong2019dropedge} is increasingly recognized for its efficacy in mitigating these shifts. The foundational approach involves learning node representations to facilitate invariant predictor learning across various constructed environments (Figure~\ref{intro} (b1)), adhering to the Risk Extrapolation (REx) principle~\cite{wu2022handling,chen2022ba,liu2023flood}.
Unfortunately, previous graph-based invariant learning methods may not effectively address the HGSS issue, primarily due to explicit environments that may be ineffective for invariant learning.
As illustrated in Figure \ref{intro} (c1), within HGSS settings, altering the original structure does not consistently affect the node's neighbor patterns. In essence, obtaining optimal and varied environments pertinent to neighbor patterns is challenging.  
Our observation (Figure \ref{intro} (c2)) reveals that EERM \cite{wu2022handling}, a pioneering invariant learning approach utilizing environment augmentation to tackle graph distribution shifts in node-level tasks,  does not perform well under HGSS settings.
At times, its enhancements are less effective than simply employing the original V-Rex~\cite{krueger2021out}, which involves randomly distributing the train nodes across various environmental groups. We attribute this phenomenon to the irrational environment construction. According to our analysis, EERM is essentially a node environment-augmented version of V-Rex, \emph{i.e.}, the disparity in their performance is solely influenced by the differing strategies in environmental construction. Besides, from the perspective of theory assumption, V-Rex is initially employed to aid model training by calculating the variance of risks introduced by different environments as a form of regularization. The significant improvements by V-Rex also reveal that the nodes of a single input heterophilic graph may reside in distinct environments, considering the variation in neighbor patterns, thus contradicting EERM's prior assumption that all nodes in a graph share the same environment~\cite{wu2022handling}. Based on this insight, our goal is to break away from previous explicit environment augmentation to learn the latent environment partition, which empowers the invariant learning to address the HGSS better.

\emph{HEI: Heterophily-Guided Environment Inference for Invariant Learning.} Recent studies explore the effect of prior knowledge on the environment partition ~\cite{lin2022zin,tan2023provably} and subsequently strengthen the importance of the environment inference and extrapolation for model generalization \cite{wu2024graph,yang2024iene}. Therefore, our initial step should be to quantify the nodes' neighbor pattern properties related to the HGSS, which is central to the issue at hand. Consequently, a critical question emerges: During the training phase, how can we identify an appropriate metric to estimate the node's neighbor pattern and leverage it to deduce latent environments to manage this HGSS issue? As previously mentioned, node homophily can assess the node's neighbor patterns \cite{lim2021large}. Unfortunately, this requires the actual labels of the node and its neighbors, rendering it inapplicable during the training stage due to the potential unlabeled status of neighbor nodes. To cope with this problem,  several evaluation metrics pertinent to nodes' neighbor patterns, including local similarity \cite{chen2023lsgnn}, post-aggregation similarity \cite{luan2022revisiting}, and SimRank~\cite{liu2023simga}, have been introduced. These metrics aim to facilitate node representation learning on heterophilic graphs during the training phase.
But these studies primarily concentrate on employing these metrics to help select proper neighbors for improved HGNN architectures, while we aim to introduce a novel invariant learning framework-agnostic backbones to separate the spurious features from selected neighbors, tackling the structure distribution shifts. 
Therefore, we propose HEI, a framework capable of generating invariant node representations through incorporating heterophily information to infer latent environments, as shown in Figure~\ref{intro} (b2), which are then used for downstream invariant prediction, under heterophilic graph structure distribution shifts. Extensive experiments on various backbones and benchmarks can verify the effectiveness of our proposed method in addressing this neglected HGSS issue.


\textbf{Our Contributions}: (i) We highlight an important yet often neglected form of heterophilic graph structure distribution shift, which is orthogonal to most HGNN works that focus on backbone designs; 
 (ii) We propose HEI, a novel graph-based invariant learning framework to tackle the HGSS issue. Unlike previous efforts, our method emphasizes leveraging a node's inherent heterophily information to deduce latent environments without augmentation, thereby significantly improving the generalization and performance of HGNNs; (iii) We demonstrate the effectiveness of HEI on several benchmarks and backbones compared with existing methods.

\section{Related Work}
\textbf{Graph Neural Networks with Heterophily.} Existing strategies for mitigating graph heterophily issues can be categorized into two groups \cite{zheng2022graph}: (i) Non-Local Neighbor Extension, aiming at identifying suitable neighbors through mixing High-order information \cite{abu2019mixhop,zhu2020beyond,jin2021universal} or discovering potential neighbors assisted by various similarity-based metrics \cite{jin2021universal,jin2021node,wang2022powerful,wang2021tree,he2022block,liu2023simga}; (ii) GNN Architecture Refinement, focusing on harnessing information derived from the identified neighbors, through selectively aggregating distinguishable and discriminative node representations, including adapting aggregation scheme \cite{luan2021heterophily,suresh2021breaking,li2022finding}, separating Ego-neighbor \cite{zhu2020beyond,suresh2021breaking,luan2021heterophily} and combining inter-layer \cite{zhu2020beyond,chen2020simple,chien2020adaptive}. However, they share the common objective of simultaneously designing better unified HGNN backbones for node classification tasks on heterophilic and homophilic graph benchmarks. Different from these works, we instead consider from an identifiable neighbor pattern distribution perspective and propose a novel invariant learning framework that can be integrated with most HGNN backbones to further enhance their performance and generalization.\\

\noindent \textbf{Generalization on GNNs.} Many efforts have been devoted to exploring the generalization ability of GNNs: (i) For graph-level tasks, it assumes that every graph can be treated as an instance for prediction tasks \cite{wu2022handling,chen2024learning}. Many works propose to identify invariant sub-graphs that decide the label Y and spurious sub-graphs related to environments, such as CIGA~\cite{chen2022learning}
, GIL~\cite{li2022learning}, GREA~\cite{liu2022graph}, DIR~\cite{wu2022discovering} and GALA \cite{chen2023does}. (ii) For node-level tasks that we focus on in this paper, the nodes are interconnected in a graph as instances in a non-iid data generation way, it is not feasible to transfer graph-level strategies directly. To address this issue, EERM \cite{wu2022handling} proposes to regard the node's ego-graph with corresponding labels as instances and assume that all nodes in a graph often share the same environment, so it should construct different environments by data augmentation, \emph{e.g.}, DropEdge \cite{rong2019dropedge}. Based on these findings, BA-GNN \cite{chen2022ba}, FLOOD \cite{liu2023flood} and IENE \cite{yang2024iene} inherit this assumption to improve model generalization. Apart from these environments-augmentation methods, SR-GNN \cite{zhu2021shift} and GDN \cite{gao2023alleviating} are two works that address distribution shifts on node-level tasks from the domain adaption and prototype learning perspectives respectively. Moreover, Renode \cite{chen2021topology} and StruRW-Mixup \cite{liu2023structural} are two reweighting-based methods that explore the effect brought by the structure difference between nodes for node classification tasks. We also compare them in experiments. Unlike these works, we highlight a special variety of structure-related distribution shifts for node classification tasks on heterophilic graphs and propose a novel invariant learning framework adapted to heterophilic graphs without dealing with graph augmentation to address this problem.

\section{Preliminaries}
\noindent \textbf{Notations}.
Given an input graph $G=(V, X, A)$, we denote $V\in \{v_1, ..., v_N\}$ as the nodes set, $X \in  R^{N\times D}$ as node features and $A\in \{0,1\}^{N \times N}$ as an adjacency matrix representing whether the nodes connect, where the $N$ and $D$ denote the number of nodes and features, respectively. The node labels can be defined as $Y\in \{0,1\}^{N \times C}$, where C represents the number of classes. For each node $v$, we use $A_{v}$ and $X_{v}$ to represent its adjacency matrix and node feature.\\

\noindent \textbf{Problem Formulation}. We first provide the formulation for the general optimized object of node-level OOD (Out of Distribution) problem on graphs, then reclarify the formulation of previous works to help distinguish our work in the next section.
From the perspective of data generation, we can get train data $\left(G_{train}, Y_{train}\right)$ from train distribution $p(\mathbf {G, Y})|\mathbf e = e)$, the model should handle the test data $\left(G_{test}, Y_{test}\right)$ from a different distribution $p(\mathbf {G, Y})|\mathbf e = e^{\prime})$, varying in different environments $\mathbf e$. Thus, the optimized object of node-level OOD problem on graphs can be formulated as follows:
\begin{small}
\begin{align}\label{eqn-ood-causal} 
\min_{\omega, \Phi} \max_{e\in \mathcal E} \mathbb E_{G}
\frac{1}{N}\sum_{v\in V}\mathbb E_{y\sim p(\mathbf y|\mathbf A_{\mathbf v}=A_v,\mathbf X_{\mathbf v}=X_v ,\mathbf e = e)} l(f_{\omega} \left(f_{\Phi}\left(A_v,X_v\right)\right), y_{v})
\end{align}
\end{small}

where the $\mathcal E$ represents the support of environments $e$, the $f_{\omega}$ and $f_{\Phi}$ refer to GNN's classifier and feature extractor respectively and $l(\cdot)$ is a loss function (e.g. cross entropy). The Eq \ref{eqn-ood-causal} aims to learn a robust model that minimizes loss across environments as much as possible. Only in this way, can the trained model be likely to adapt to unknown target test distribution well. However, environmental labels for nodes are usually unavailable during the training stage, which inspires many works to seek methods to make use of environmental information to help model training.\\

\noindent \textbf{Previous Graph-based Invariant Learning.} To approximate the optimized object of Eq. \ref{eqn-ood-causal}, previous works \cite{wu2022handling,chen2022ba,liu2023flood} mainly construct diverse environments by adopting the masking strategy as Figure \ref{intro} (b1).
Thus, we conclude previous works from the masking strategy ($Mask_{\eta}(\cdot)$ parameterized with $\eta$). Given an input single graph, we can obtain K augmented graphs as Eq. \ref{mask}, where each graph corresponds to an environment. The K is a pre-defined number of training environments and $X^{m}$, $A^{m}$, and $V^{m}$ are corresponding mask versions of feature, adjacency matrix, and node sets. 
\begin{small}
\begin{equation}\label{mask}
G^{e=k} = Mask^{e=k}_{\eta}(G) = \left(X^{m},A^{m},V^{m}\right)_{e=k}, k=1,2...,K
\end{equation}
\end{small}

Then, assisted by these augmented graphs with environment labels, the GNN $f(\cdot)$ parameterized by $\left(\omega,\Phi\right)$ can be trained considering environmental information. We can define the ERM (Empirical Risk Minimization) loss in the $k$-th environment as the Eq. \ref{EERM_base}, which only calculates the loss on the corresponding augmented $G^{e=k}$.

\begin{small}
\begin{equation}\label{EERM_base}
    R_{e=k}\left(\omega, \Phi \right) =\frac{1}{N}\sum_{v\in V^{m}} l\left(f_\omega(f_\Phi\left(A_v,X_v\right), y_{v}\right)
\end{equation}
\end{small}

Following the principle of Variance Risk Extrapolation (V-Rex) to reduce the risks from different environments, the final training framework can be defined as Eq. \ref{EERM}. Where the $\lambda$ controls the effect between reducing the average risk and promoting equality of risks. 
\begin{small}
\begin{align}\label{EERM}
&\min_{\omega, \Phi} \max_{\eta} ~ L\left(\Phi, \omega, \eta \right) = \notag\\
&\sum\nolimits_{k=1}^K{R_{e=k}(\omega, \Phi)} + \lambda Var \left(R_{e=1}(\omega, \Phi),\cdots ,R_{e=K}(\omega, \Phi)\right)
\end{align}
\end{small}

The maximization means that we should optimize the masking strategy (parameter $\eta$) to construct sufficient and diverse environments, while the minimization aims to reduce the training loss for the model (parameter $\omega$ and $\Phi$).\\

\noindent \textbf{Discussions.} Exactly, previous graph-based invariant learning methods introduce extra augmented graphs to construct nodes' environments while our work only infers nodes' environments on a single input graph. Specifically, there exists a latent assumption for previous works that nodes on a single graph belong to the same environment so we need to construct diverse environments by data augmentation. This assumption arises from the insight that nodes on an input graph come from the same \emph{outer domain-related environments} (e.g. Financial graphs or Molecular graphs) \cite{wu2022handling}. But considering the message-passing mechanism on heterophilic graphs (the ideal aggregation target should be nodes with the same label), the nodes should exist exactly in \emph{inner structure-related environments}. To cope with this issue, as shown in Figure \ref{intro} (c1), directly utilizing \emph{data augmentation may be ineffective in changing the node's neighbor pattern distribution} to construct diverse environments for invariant prediction. At the same time, the neighbor pattern difference between train and test has verified that even on a single graph, the nodes may belong to different structure-related environments. These simultaneously inspire us to directly infer node environments on a single graph assisted by the node's neighbor pattern, rather than constructing environments from different augmented graphs, for addressing heterophilic graph structure distribution shift.

\begin{figure*}
\begin{center}
\centering
\setlength{\abovecaptionskip}{0cm}   
\setlength{\belowcaptionskip}{0cm}   
\includegraphics[width=0.8\linewidth]{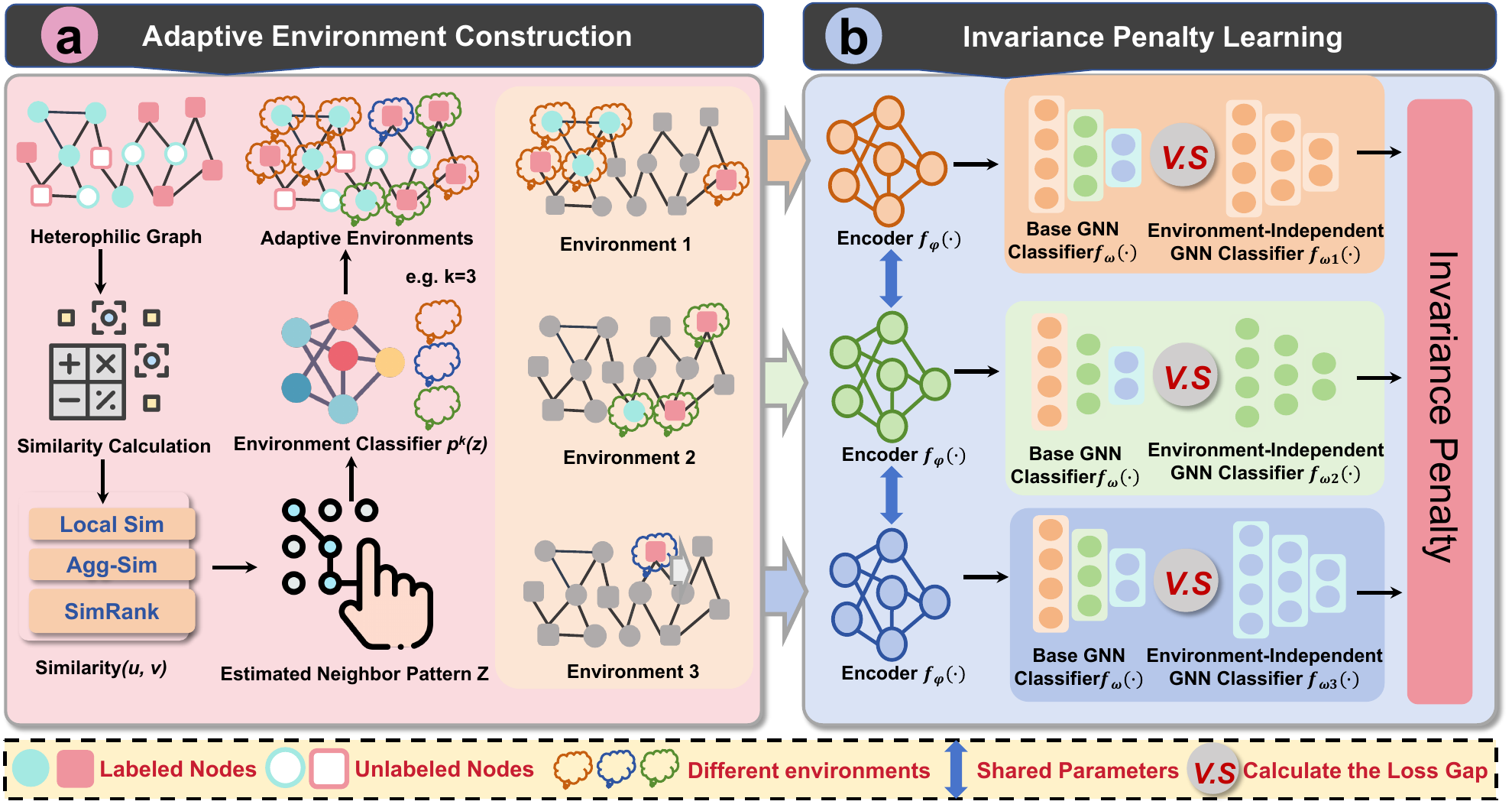}
\caption{Illustrations of our framework HEI. (a) The neighbor pattern for each train node can be estimated by similarity first and then used for inferring environments without augmentation; (b) Based on the train nodes belonging to different inferred environments, we can train a set of environment-independent GNN classifiers with the shared encoder compared with the base GNN. The shared encoder outputs the representations of nodes in each environment and then forwards them to the base GNN classifier and the environment-independent classifier respectively. By calculating the loss gap between these two different classifiers, an invariance penalty is introduced to improve model generalization.}
\Description{Illustrations of our framework HEI}
\label{method}
\end{center}
\vspace{-0.3cm}  
\end{figure*}
\section{Methodology}\label{section_method}
In this section, we present the details of the proposed HEI. Firstly, on heterophilic graphs, we verify that the similarity can serve as a neighbor pattern indicator and then review existing similarity-based metrics to estimate the neighbor patterns during training stages. Then, we elaborate the framework to jointly learn environment partition and invariant node representation on heterophilic graphs without augmentation, assisted by the estimated neighbor patterns. Finally, we clarify the overall training process of the algorithm and discuss its complexity. 


\noindent \subsection{Neighbor Patterns Estimation}
The node homophily is commonly used to evaluate the node's neighbor patterns, representing the node's structural pattern distribution \cite{lim2021large}. Unfortunately, it needs the true labels of the node and its neighbors, which means they can not be used in the training stage because the neighbor nodes may be just the test nodes without labels for node classification tasks when given an input graph. To cope with it, we aim to utilize the similarity between node features to estimate the node's neighbor pattern.\\


\noindent \textbf{Similarity: An Indicator of Neighbor Patterns.} 
Previous works have shown there exists a certain relationship between similarity and homophily from the experimental analysis \cite{chen2023lsgnn}, it can not be guaranteed to work well without a theory foundation. Thus, we further investigate its effectiveness from the node cluster view and verify the similarity between nodes can be exploited to approximate the neighbor pattern without the involvement of label information.

For simplicity, we take K-Means as the cluster algorithm. For two nodes $v$ and $u$, let $v$ belong to the cluster centroid $c_{1}$ and denote the square of the distance between $v$ and $u$ as $\delta = \left\|u-v\right\|^{2}$, we can get $c_{1}={\arg \min }\left\|v-c_{i}\right\|^{2}$, where $c_{i}$ represents the $i$-th cluster centroid. Then the distance between $u$ and cluster centroid $c_{1}$ can be acquired as the Eq. \ref{cluster}. Exactly, the neighbor pattern describes the label relationship between the node and its neighbors. From the Eq \ref{cluster}, we can find the smaller $\delta$, the more likely the $v$ and $u$ belong to the same cluster and own the same label. Therefore, the similarity between nodes can be exploited to serve as a neighbor pattern indicator without using label information.
\begin{small}
\begin{align}
\begin{array}{l}
\left\|u-c_{1}\right\|^{2}=\left\|\left(u-v\right)+\left(v-c_{1}\right)\right\|^{2} \\
=\left\|\left(u-v\right)\right\|^{2}+2\left\|u-v\right\|\left\|v-c_{1}\right\|+\left\|v-c_{1}\right\|^{2} \\
=\delta+2 \sqrt{\delta}\left\|v-c_{1}\right\|+\left\|v-c_{1}\right\|^{2} \\
=\left(\left\|v-c_{1}\right\|+\sqrt{\delta}\right)^{2} \geq \delta
\end{array}\label{cluster}
\end{align}
\end{small}

\noindent \textbf{Existing Similarity-based Metrics.} Existing similarity-based metrics on heterophilic graphs can be shown as Eq.\ref{definitions}. 

\begin{small}
\begin{equation}\label{definitions}
\text{Similarity}(u,v) = 
\begin{cases}
    \operatorname{Sim}(X_{v}, X_{u}) & \text{Local Sim} \\ 
    \operatorname{Sim}(\hat{A_{v}}X_{v}, \hat{A_{u}}X_{u}) & \text{Agg Sim} \\ 
    \frac{c}{|NS(u)||NS(v)|}\sum\limits_{\substack{u^{\prime}\in NS(u) \\ v^{\prime}\in NS(v)}} \operatorname{Sim}(X_{u^{\prime}},X_{v^{\prime}}) & \text{SimRank}
\end{cases}
\end{equation}
\end{small}
Let \( c \in (0, 1) \) be a decay factor set to \( c = 0.6 \), \( \mathcal{N}(v) \) denote the neighbor set of node \( v \), \( \hat{A}_v \) represent the aggregation operation on \( v \), and \( \text{Sim} \) measure the similarity between two objects. Local Sim~\cite{chen2023lsgnn} and Agg-Sim~\cite{luan2022revisiting} compute the similarity between original and post-aggregation embeddings, while SimRank~\cite{liu2023simga} measures similarity based on the neighbor nodes of nodes.\\


\noindent \textbf{Estimated Node's Neighbor Pattern.} Thus, as Eq.\ref{pattern}, we can obtain the estimated neighbor patterns $z_{v}$ for the node $v$ during the training stage by averaging the node's similarity with neighbors.
\begin{small}
\begin{equation} \label{pattern}
z_{v}= \frac{1}{|NS(v)|} \sum_{u\in NS(v)} Similarity(u,v)
\end{equation}    
\end{small}

Notably, we further strengthen our object of using similarity metrics is indeed different from previous HGNN works that utilize the similarity metrics \cite{luan2022revisiting,chen2023lsgnn,liu2023simga} to design backbones. From the perspective of causal analysis, when given the neighbors, we aim to separate and weaken the effect of spurious features from full neighbor features by utilizing the estimated neighbor pattern to infer the node's environment for invariant prediction. However, previous HGNN works mainly aim to help the node select proper neighbors and then directly utilize full neighbor features as aggregation targets for better HGNN backbone designs. Our work is exactly \emph{orthogonal} to previous HGNN works.

\subsection{HEI: Heterophily-Guided Environment Inference for Invariant Learning} 


We aim to use the estimated neighbor patterns $Z \in \mathbb{R}^{G_z}$, which represent node heterophily, as an auxiliary tool to jointly learn environment partition and invariant node representation without augmentation, similar to techniques in \cite{chang2020invariant, lin2022zin} for image classification. With the estimated neighbor patterns, we train an environment classifier $\rho(\cdot): \mathbb{R}^{G_z} \rightarrow \mathbb{R}^K$ to assign nodes to $K$ environments, where $\rho$ is a two-layer MLP and $\rho^{(k)}(\cdot)$ denotes the $k$-th entry of $\rho(\cdot)$. We have $\rho(Z) \in [0, 1]^K$ and $\sum_k \rho^{(k)}(Z) = 1$. The ERM loss for all train nodes is denoted as $R(\omega, \Phi)$, and the ERM loss for the $k$-th inferred environment is defined as $R_{\rho^{(k)}}(\omega, \Phi)$, calculated only on nodes in the $k$-th environment.\\

\noindent \textbf{Overal Framework:} Based on the above analysis, the training framework of HEI can be defined as Eq.\ref{erm_enviroments} and Eq.\ref{penalty}. 


\begin{small}
\begin{align}
\label{erm_enviroments}
R_{\rho^{(k)}}(\omega, \Phi) =\frac{1}{N}\sum_{v\in V} \rho^{(k)}(z_v)l \left(f_\omega(f_\Phi\left(A_v,X_v\right), y_{v}\right) 
\end{align}
\end{small}

\begin{small}
\begin{align}
&\min_{\omega, \Phi} \; \max_{ \rho, \{\omega_1,\cdots,\omega_K\}}~L(\Phi, \omega, \omega_1,\cdots,\omega_K, \rho)  = \notag \\ 
&R(\omega, \Phi)+  \lambda\underbrace {\sum \nolimits_{k=1}^K\big[R_{\rho^{(k)}}(\omega, \Phi) - R_{\rho^{(k)}}(\omega_k, \Phi)\big]}_{\text{\small invariance penalty}}
\label{penalty}
\end{align}
\end{small}

Compared with previous graph-based invarinat learning methods shown in Eq. \ref{EERM_base} and Eq. \ref{EERM}, our framework mainly differs in the maximization process. Thus, we clarify the effectiveness and reasonability of our framework from two aspects: (i) The invariance penalty learning that introduces a set of environment-dependent GNN classifiers $\{f_{\omega_k}\}_{k=1}^K$, which are only trained on the data belonging to the inferred environments; (ii) The adaptive environment construction through optimizing the environmental classifier $\rho(\cdot)$.\\

\noindent \textbf{Invariance Penalty Learning.} As shown by Eq.\ref{eqn-ood-causal}, the ideal GNN classifier $f_\omega$ is expected to be optimal across all environments. After the environment classifier $\rho^{(k)}(\cdot)$ assigns the train nodes into k inferred environments, we can adopt the following criterion to check if $f_\omega$ is already optimal in all inferred environments: Take the $k$-th environment as an example, we can additionally train an environment-dependent classifier $f_{\omega_k}$ on the train nodes belonging to the $k$-th environment.  If $f_{\omega_k}$ achieves a smaller loss, it indicates that $f_\omega$ is not optimal in this environment. We can further train a set of classifiers $\{f_{\omega_k}\}_{k=1}^K$, each one with a respective individual environment, to assess whether $f_\omega$ is simultaneously optimal in all environments. Notably, all these classifiers share the same encoder $f_{\Phi}$, if $f_{\Phi}$ extracts spurious features that are unstable across the inferred environments, $R_{\rho^{(k)}}(\omega, \Phi)$ will be larger than $R_{\rho^{(k)}}(\omega_k, \Phi)$, resulting in a non-zero invariance penalty, influencing model optimization towards achieving optimality across all environments. As long as the encoder extracts the invariant feature, the GNN classifier $f_{\omega}$ and its related environment-dependent classifier $\{f_{\omega_k}\}_{k=1}^K$ will have the same prediction across different environments. \\



\noindent \textbf{Adaptive Environment Construction.} As shown in Figure \ref{intro} (c), the effectiveness of previous methods is only influenced by environmental construction strategy. A natural question arises: What is the ideal environment partition for invariant learning to deal with the HGSS? We investigate it from the optimization of environment classifier $\rho(\cdot)$. Specifically, a good environment partition should construct environments where the spurious features exhibit instability, incurring a large penalty if $f_{\Phi}$ extracts spurious features. In this case, we should maximize the invariance penalty to optimize the partition function $\rho(\cdot)$ to generate better environments, which is also consistent with the proposed strategy. Though previous works \cite{wu2022handling,chen2022ba,liu2023flood} also adopt the maximization process to construct diverse environments, they just focus on directly optimizing the masking strategy to get augmentation graphs. During the optimization process, these methods lack guidance brought by auxiliary information $Z$ related to environments, ideal or effective environments are often unavailable in this case. That's why we propose to introduce the environment classifier to infer environments without augmentation, assisted by the $Z$. Exactly, to make sure the guidance of $Z$ has a positive impact on constructing diverse and effective environments for the invariant node representation learning, there are also two conditions for $Z$ from the causal perspective.

 \begin{table*}
\setlength{\abovecaptionskip}{0cm}
\setlength{\belowcaptionskip}{0cm}
\captionsetup{font={small,stretch=1.25}, labelfont={bf}}
 \renewcommand{\arraystretch}{1}
    \caption{Performance comparison on small-scall heterophilic graph datasets under Standard Settings. The reported scores denote the classification accuracy (\%) and error bar (±) over 10 trials.  We highlight the best score on each dataset in bold and the second score with underline.}
    \centering
    \resizebox{0.9\textwidth}{!}{
        \begin{tabular}{ll|c|c|c|c|c|c|c|c|c|c}
            \toprule[1.5pt]
            {}&\textbf{\multirow{2}*{Backbones}} &\textbf{\multirow{2}*{Methods}} &\multicolumn{3}{|c|}{Chamelon-filter} 
            &\multicolumn{3}{|c|}{Squirrel-filter} &\multicolumn{3}{c}{Actor} \\\cline{4-12} 
            \textbf{} & \textbf{} &\textbf{} &Full Test  &High Hom Test &Low Hom Test &Full Test  &High Hom Test & Low Hom Test&Full Test  &High Hom Test &Low Hom Test\\ 
            \toprule[1.0pt]
            &\multirow{11}*{LINKX} 
            & ERM   &23.74 ± 3.27 &29.24 ± 3.87  &16.81 ± 3.21 &37.11 ± 1.68 &45.11 ± 1.68 &28.17 ± 1.54&35.83 ± 1.40 &38.21 ± 1.76&33.42 ± 1.94 \\
            &\textbf{} &{ReNode} &24.14 ± 3.51 &29.64 ± 3.42  &17.24 ± 3.25 &37.91 ± 1.84 &45.81 ± 1.58 &28.97 ± 1.24&28.62 ± 1.79&33.92 ± 3.67&23.25 ± 1.54\\

            &\textbf{} &{StruRW-Mixup} &24.19 ± 3.81 &29.84 ± 3.58  &17.46 ± 3.27 &37.98 ± 1.57 &45.88 ± 1.81 &29.27 ± 1.24&28.62 ± 1.79&33.92 ± 3.67&23.25 ± 1.54  \\
                        
            &\textbf{} &{SRGNN}&24.42 ± 3.57 &29.94 ± 3.59  &17.96 ± 3.12 &38.11 ± 1.67 &45.98 ± 1.82 &29.57 ± 1.23 &29.57 ± 1.81&34.45 ± 3.51&24.67 ± 1.34\\
            
            &\textbf{}&{EERM}&24.54 ± 3.61 &30.12 ± 3.71  &18.22 ± 3.42 &38.57 ± 1.77 &46.24 ± 1.82 &29.57 ± 1.23&29.62 ± 1.94&36.50 ± 3.21&22.66 ± 1.26\\

            &\textbf{}&{BAGNN}&24.64 ± 3.71 &30.17 ± 3.61  &18.29 ± 3.54 &38.61 ± 1.72 &46.14 ± 1.93 &29.87 ± 1.15&36.10 ± 2.01&38.46 ± 2.16&33.49 ± 1.86\\
            
            &\textbf{}&{FLOOD}&24.64 ± 3.62 &30.15 ± 3.62  &18.25 ± 3.82 &38.59 ± 1.81 &46.33 ± 1.52 &29.64 ± 1.13 &36.40 ± 2.31&38.72 ± 2.17&33.98 ± 1.76\\

            &\textbf{}&{CaNet}&24.71 ± 3.55 &30.28 ± 3.28  &18.45 ± 3.78 &38.75 ± 1.85 &46.45 ± 1.55 &29.74 ± 1.34 &36.58 ± 2.31& 38.92 ± 2.44&34.12 ± 1.48\\

            &\textbf{}&{IENE}&\underline{24.74 ± 3.58} &30.30 ± 3.62  &\underline{18.85 ± 3.82} &38.81 ± 1.86 &46.43 ± 1.59 &29.84 ± 1.75 &\underline{36.65 ± 2.31}& \underline{38.99 ± 2.35}&\underline{34.13 ± 1.96}\\
                      
            &\textbf{}&{GDN}&24.65 ± 3.52 &\underline{30.31 ± 3.62}  &18.15 ± 3.82 &\underline{38.89 ± 1.81} &\underline{46.58 ± 1.57} &\underline{29.94 ± 1.87} &36.38 ± 2.25&38.52 ± 2.17&34.08 ± 1.76\\
            
            &\textbf{}&\textbf{HEI(Ours)}&\textbf{25.78 ± 2.23}	&\textbf{31.35 ± 2.56}	&\textbf{20.21 ± 4.21}& \textbf{40.92 ± 1.31}	&\textbf{48.82 ± 2.88}	&\textbf{31.21 ± 2.79} & \textbf{37.41 ± 1.17}	&\textbf{39.31 ± 1.45}&\textbf{35.42 ± 1.54}\\
            \toprule[1.0pt]
            &\multirow{11}*{GloGNN++}
            &ERM  &25.90 ± 3.58 & 31.40 ± 3.63 &20.10 ± 2.51 &35.11 ± 1.24 &41.25 ± 1.88 &27.61 ± 1.54  &37.70± 1.40 & 40.96 ± 1.52 & 34.37 ± 1.61\\ 
            &\textbf{}&{ReNode}&25.99 ± 3.88 & 31.50 ± 3.83 &20.20 ± 2.32 &35.54 ± 1.12 &41.54 ± 1.78 &28.21 ± 1.59 &29.82 ± 1.79&35.42 ± 2.75&25.24 ± 2.18\\
            &\textbf{}&{StruRW-Mixup}&26.12 ± 3.58 & 31.70 ± 3.53 &20.30 ± 2.12 &35.64 ± 1.18 &41.70 ± 1.58 &28.31 ± 1.56 &29.87 ± 1.81&35.45 ± 2.79&25.34 ± 2.51\\        
            
            &\textbf{} &{SRGNN}&26.72 ± 3.68 & 32.20 ± 3.43 &21.00 ± 2.52 &36.34 ± 1.28 &42.30 ± 1.58 &28.75 ± 1.54&30.87 ± 1.79&35.62 ± 2.75&25.64 ± 2.54\\
            
            &\textbf{}&{EERM}&26.99 ± 3.58 & 32.51 ± 3.73 &21.22 ± 2.41 &36.54 ± 1.38 &42.70 ± 1.48 &29.45 ± 1.55 &32.75 ± 2.41&39.34 ± 3.21&26.98 ± 2.87\\
            
            &\textbf{}&{BAGNN}&27.12 ± 3.48 & 32.61 ± 3.83 &21.42 ± 2.71 &36.64 ± 1.52 &42.89 ± 1.49 &29.55 ± 1.61 &38.05 ± 1.29 &41.26 ± 1.52 & 34.87 ± 1.87\\
            
            &\textbf{}&{FLOOD}&27.17 ± 3.58 & 32.81 ± 3.63 &21.82 ± 2.61 &36.84 ± 1.42 &42.97 ± 1.58 &29.85 ± 1.57 &38.35 ± 1.59&41.54 ± 1.34 & 34.99 ± 2.11\\

            &\textbf{}&{CaNet}&27.37 ± 3.27 & 32.99 ± 3.87&22.08 ± 2.51 &36.99 ± 1.51 &43.17 ± 1.66 &30.15 ± 1.44 &38.37 ± 1.61&41.58 ± 1.32 & 35.01± 2.10\\

            &\textbf{}&{IENE}&\underline{27.75 ± 3.31} & \underline{33.35 ± 3.97}&\underline{22.62 ± 2.49} &\underline{37.15 ± 1.66} &\underline{43.37 ± 1.66} &\underline{30.38 ± 1.64} &38.38 ± 1.66&41.58 ± 1.37 & 35.08 ± 2.12\\

            &\textbf{}&{GDN}&27.21 ± 3.68 & 32.91 ± 3.52&21.84 ± 2.21 &36.66 ± 1.42 &43.12 ± 1.57 &29.65 ± 1.58 &\underline{38.39 ± 1.69}&\underline{41.59 ± 1.74} & \underline{35.12 ± 2.51}\\
            
            &\textbf{}&\textbf{HEI(Ours)}
            &\textbf{29.31 ± 3.68} & \textbf{34.35 ± 3.52} &\textbf{24.25± 2.71} &\textbf{39.42 ± 1.51} &\textbf{45.19 ± 1.57} &\textbf{31.45 ± 1.68}&\textbf{39.41 ± 1.51} & \textbf{42.25 ± 1.59} & \textbf{36.12 ± 1.85}\\
            \toprule[1.5pt]    
        \end{tabular}
    }
    \label{small_zin}
    \vspace{-0.3cm}
\end{table*}

\begin{table*}
\setlength{\abovecaptionskip}{0cm}
\setlength{\belowcaptionskip}{0cm}
\captionsetup{font={small,stretch=1.25}, labelfont={bf}}
 \renewcommand{\arraystretch}{1}
    \caption{Performance comparison on large-scall heterophilic graph datasets under Standard Settings. The reported scores denote the classification accuracy (\%) and error bar (±) over 5 trials. We highlight the best score on each dataset in bold and the second score with underline. }
    \centering
    \resizebox{0.9\textwidth}{!}{
        \begin{tabular}{ll|c|c|c|c|c|c|c|c|c|c}
            \toprule[1.5pt]
            {}&\textbf{\multirow{2}*{Backbones}} &\textbf{\multirow{2}*{Methods}} &\multicolumn{3}{|c|}{Penn94} 
            &\multicolumn{3}{|c|}{arxiv-year} &\multicolumn{3}{c}{twitch-gamer} \\\cline{4-12} 
            \textbf{} & \textbf{} &\textbf{} &Full Test  &High Hom Test &Low Hom Test &Full Test  &High Hom Test & Low Hom Test&Full Test  &High Hom Test &Low Hom Test\\ 
            \toprule[1.0pt]
            &\multirow{11}*{LINKX} &ERM   &84.67 ± 0.50&87.95 ± 0.73 &81.07 ± 0.50 &54.44 ± 0.20 &64.74 ± 0.42&\underline{48.39 ± 0.62} &66.02 ± 0.20&\underline{85.47 ± 0.66}&46.38 ± 0.67\\
            &\textbf{}&{ReNode}&84.91 ± 0.41&88.02 ± 0.79&81.53 ± 0.88&54.46 ± 0.21&64.80 ± 0.37&48.37 ± 0.55&66.13 ± 0.14&84.25 ± 0.48&47.84 ± 0.43\\
            &\textbf{}&{StruRW-Mixup}&84.96 ± 0.43&88.11 ± 0.56&81.91 ± 0.78&54.35 ± 0.21&64.78 ± 0.32&48.31 ± 0.81&66.10 ± 0.12&84.29 ± 0.47&47.89 ± 0.58\\
            &\textbf{} &{SRGNN}&84.98 ± 0.37&87.92 ± 0.79&81.83 ± 0.78&54.42 ± 0.20&64.80 ± 0.37&48.38 ± 0.54&66.15 ± 0.09&84.45 ± 0.48&48.01 ± 0.43\\
            &\textbf{}&{EERM}&85.01 ± 0.55&87.81 ± 0.79&82.08 ± 0.71&\underline{54.82 ± 0.32}&\textbf{68.06 ± 0.61}&46.46 ± 0.61&66.02 ± 0.18&83.27 ± 0.40&48.39 ± 0.34\\
            &\textbf{}&{BAGNN}&85.02 ± 0.37&88.21 ± 0.68&82.02 ± 0.59&54.65 ± 0.30&66.46 ± 0.57&47.56 ± 0.58&66.17 ± 0.12&83.77 ± 0.40&48.56 ± 0.59\\
            &\textbf{}&{FLOOD}&85.07 ± 0.32&88.25 ± 0.59&82.11 ± 0.61&54.77 ± 0.29&66.81 ± 0.59&47.88 ± 0.56&66.18 ± 0.14&83.85 ± 0.42&48.71 ± 0.61\\
            
            &\textbf{}&{CaNet}&85.10 ± 0.28&88.33 ± 0.54&82.15 ± 0.61&54.78 ± 0.29&66.88 ± 0.57&47.91 ± 0.66&66.20 ± 0.15&83.95 ± 0.42&48.78 ± 0.61\\

            &\textbf{}&{IENE}&85.15 ± 0.32&88.25 ± 0.59&82.20 ± 0.60&54.80 ± 0.33&66.85 ± 0.61&48.01 ± 0.59&\underline{66.21 ± 0.12}&84.45 ± 0.47&\underline{48.81 ± 0.61}\\

            &\textbf{}&{GDN}&\underline{85.19 ± 0.37}&\underline{88.31 ± 0.68}&\underline{82.32 ± 0.59}&54.75 ± 0.30&\underline{66.96 ± 0.61}&47.86 ± 0.58&66.12 ± 0.12&83.77 ± 0.40&48.56 ± 0.59\\
            &\textbf{}&\textbf{HEI(Ours)}&\textbf{86.22 ± 0.28}	&\textbf{89.24 ± 0.28}&\textbf{83.22 ± 0.59} &\textbf{56.05 ± 0.22}&	66.53 ± 0.41	&\textbf{49.33 ± 0.32} &\textbf{66.79 ± 0.14} & \textbf{85.53 ± 0.25} &\textbf{49.21 ± 0.57} \\
            \toprule[1.0pt]

            &\multirow{11}*{GloGNN++} &ERM  &85.81 ± 0.43 &89.51 ± 0.82 & 81.75 ± 0.58 &54.72 ± 0.27 &65.78 ± 0.41 &48.12 ± 0.72  &66.29 ± 0.20&84.25 ± 1.06&48.13 ± 1.06\\ 
            &\textbf{}&{Renode}&85.81 ± 0.42&89.53 ± 0.81&81.75 ± 0.57&54.76 ± 0.25&65.91 ± 0.54&48.12 ± 0.75&66.32 ± 0.16&84.01 ± 0.56&48.44 ± 0.67\\
            &\textbf{}&{StruRW-Mixup}&85.92 ± 0.37&89.83 ± 0.81&81.81 ± 0.47&54.81 ± 0.35&65.98 ± 0.64&48.52 ± 0.65&66.29 ± 0.15&84.21 ± 0.56&48.54 ± 0.67\\
            &\textbf{} &{SRGNN}&85.89 ± 0.42&89.63 ± 0.81&82.01 ± 0.57&54.69 ± 0.25&65.87 ± 0.44&48.39 ± 0.85&66.25 ± 0.16&\underline{84.31 ± 0.56}&48.34 ± 0.57\\
            &\textbf{}&{EERM}&85.86 ± 0.33 &89.41 ± 0.74&81.97 ± 0.50&53.11 ± 0.19&61.03 ± 0.54&48.54 ± 0.43&66.20 ± 0.30&83.97 ± 1.18&48.25 ± 0.96\\
            &\textbf{}&{BAGNN}&85.95 ± 0.27 &89.61 ± 0.74&81.92 ± 0.40 &54.81 ± 0.17&66.07 ± 0.44&48.39 ± 0.33&66.22 ± 0.25&83.77 ± 0.85&\underline{48.64 ± 0.59}\\
            &\textbf{}&{FLOOD}&85.99 ± 0.31 & 89.64 ± 0.67&82.05 ± 0.51 &54.89 ± 0.22&66.22 ± 0.42&48.55 ± 0.31&66.24 ± 0.22&83.81 ± 0.79&48.50 ± 0.54\\

            &\textbf{}&{CaNet}&86.02 ± 0.35 & \underline{89.69 ± 0.70}&82.11 ± 0.57 &\underline{54.91 ± 0.33}&\underline{66.28 ± 0.42}&\underline{48.68 ± 0.33}&66.28 ± 0.27&83.89 ± 0.78&48.62 ± 0.59\\

            &\textbf{}&{IENE}&\underline{86.11 ± 0.30} & 89.64 ± 0.68&\underline{82.18 ± 0.47} &54.86 ± 0.19&66.21 ± 0.44&48.65 ± 0.33&\underline{66.35 ± 0.25}&84.11 ± 0.78&\underline{48.79 ± 0.44}\\

            &\textbf{}&{GDN}&85.92 ± 0.41 & 89.53 ± 0.67&81.75 ± 0.51 &54.76 ± 0.17&66.24 ± 0.44&48.15 ± 0.53&66.21 ± 0.27&83.78 ± 0.89&48.42 ± 0.51\\
            
            &\textbf{}&\textbf{HEI(Ours)} &\textbf{87.18 ± 0.28}	&\textbf{89.99 ± 0.65}	&\textbf{83.59 ± 0.39} &\textbf{55.71 ± 0.24}&\textbf{66.29 ± 1.14}	&\textbf{49.52 ± 0.75} 
            &\textbf{66.99 ± 0.17} &\textbf{84.37 ± 0.68}&\textbf{50.40 ± 0.52}
            \\
            \toprule[1.5pt]    
        \end{tabular}
    }
    \label{large_zin}
    \vspace{-0.2cm}
\end{table*}
\section{Experiments}
In this section, we investigate the effectiveness of HEI
to answer the following questions.

\begin{itemize}
    \item \textbf{RQ1:} Does HEI outperform state-of-art methods to address the HGSS issue?
    \item \textbf{RQ2:} How robust is the proposed method? Can HEI solve the problem that exists in severe distribution shifts? 
    \item \textbf{RQ3:} How do different similarity-based metrics influence the neighbor pattern estimation, so as to further influence the effect of HEI?
   \item \textbf{RQ4:} What is the sensitivity of HEI concerning the pre-defined number of training environments? 
\end{itemize}

\subsection{Experimental Setup} \label{set}
\textbf{Datasets}. We adopt six commonly used heterophilic graph datasets (chameleon, Squirrel, Actor, Penn94, arxiv-year, and twitch-gamer) to verify the effectiveness of HEI \cite{pei2020geom,lim2021large}. 
To make sure the evaluation is stable and reasonable, we utilize the filtered versions of existing datasets to avoid data leakage \cite{platonov2023critical}. Notably, considering that we should further split the test datasets to construct different evaluation settings. Those excessive small-scale heterophilic graph datasets, such as Texa, Cornell, and Wisconsin \cite{pei2020geom}, are not fit and chosen for evaluation due to their unstable outcomes.\\ 


\noindent \textbf{Settings.} Based on previous dataset splits, we construct two different settings to evaluate the effectiveness and robustness of HEI: \textbf{(i) Standard Settings:} We sort the test nodes based on their nodes' homophily values and acquire the median. The part that is higher than the median is defined as the High Hom Test, while the rest is defined as the Low Hom Test. The model is trained on the previous train dataset and evaluated on more fine-grained test groups; \textbf{(ii) Simulation Settings where exists severe distribution shifts.:} We sort and split the train and test nodes simultaneously adopting the same strategy of (i). The model is trained on the Low/High Hom Train and evaluated on the High/Low Hom Test.\\

\noindent \textbf{Backbones.} To further verify our framework is orthogonal to previous HGNN works that focus on backbone designs, we adapt HEI to two existing SOTA and scalable backbones with different foundations, LINKX (MLP-based) \cite{lim2021large} and GloGNN++ (GNN-based) \cite{li2022finding}. In this way, our improvements can be attributed to the design that deals with the neglected heterophilic structure distribution shifts.\\

\noindent \textbf{Baselines}. Denote the results of the backbone itself as ERM. Our comparable baselines can be categorized into: (i) Reweighting-based methods considering structure information: Renode \cite{chen2021topology} and StruRW-Mixup \cite{liu2023structural}; (ii) Invariant Learning methods involving environment inference for node-level distribution shift: SRGNN \cite{zhu2021shift}, EERM \cite{wu2022handling}, BAGNN \cite{chen2022ba}, FLOOD \cite{liu2023flood}, CaNet \cite{wu2024graph} and IENE \cite{yang2024iene} ; (iii) Prototype-based methods for structural distribution shift on the special domain (e.g. graph anomaly detection): GDN \cite{gao2023alleviating}.
Notably, though we can utilize estimated neighbor patterns as auxiliary information to infer environments related to HGSS, the true environment label is still unavailable. So we don't compare with those traditional invariant learning methods that rely on the explicit environment labels, e.g. IRM \cite{arjovsky2019invariant}, V-Rex \cite{krueger2021out} and GroupDRO \cite{sagawa2019distributionally}.
\subsection{Experimental Results and Analysis}\label{experiments}
\noindent \textbf{Handling Distribution Shifts under Standard Settings (RQ1).}
We first evaluate the effectiveness of HEI  under standard settings, where we follow the previous dataset splits and further evaluate the model on more fine-grained test groups with low and high homophily, respectively. The results can be shown in Table \ref{small_zin} and Table \ref{large_zin}. We have the following observations.

On the one hand, \emph{the impact brought by the HGSS is still apparent though we adopted the existing SOTA HGNN backbones.} As shown by the results of ERM in Table \ref{small_zin} and Table \ref{large_zin}, for most datasets, there are significant performance gaps between the High Hom Test and Low Hom Test, ranging from 5 to 30 scores. These results further verify the necessity to seek methods from the perspective of data distribution rather than backbone designs to deal with this problem. 

On the other hand, \emph{HEI can outperform previous methods in most circumstances.} Specifically, compared with invariant learning methods, though HEI does not augment the training environments, utilizing the estimated neighbor patterns to directly infer latent environments still benefits invariant prediction and improves model generalization on different test distributions related to homophily. In contrast,  directly adopting reweighting-based strategies (Renode and StruRW) or evaluating the difference between the training domain and target domain (SRGNN) without environment augmentation can't acquire superior results than invariant learning methods. This is because these methods need accurate domain knowledge or structure information in advance to help model training. However, for the HGSS issue, the nodes' environments on heterophily graphs are unknown and difficult to split into the invariant and spurious domains, like the GOOD dataset \cite{gui2022good}, which has clear domain and distribution splits. Simultaneously, the neighbor pattern distribution represents more fine-grained label relationships between nodes and their neighbors, which means it's more complex and challenging compared to previous long-tail degree or label problems depending on directly counting class types and numbers of neighbors. Moreover, the great performance from GDN verifies the necessity of learning node representation close to its class prototype through regularization, mitigating the effect of neighbor patterns during aggregation. However, HEI can still outperform GDN due to the more fine-grained causal feature separation depending on the constructed environments related to the node's neighbor patterns, further verifying the effectiveness of HEI. \\


\begin{figure*}
\setlength{\abovecaptionskip}{0cm}   
\setlength{\belowcaptionskip}{0cm}   
\begin{center}
\subfigure[Chameleon-filter]{
\includegraphics[width=0.25\linewidth]{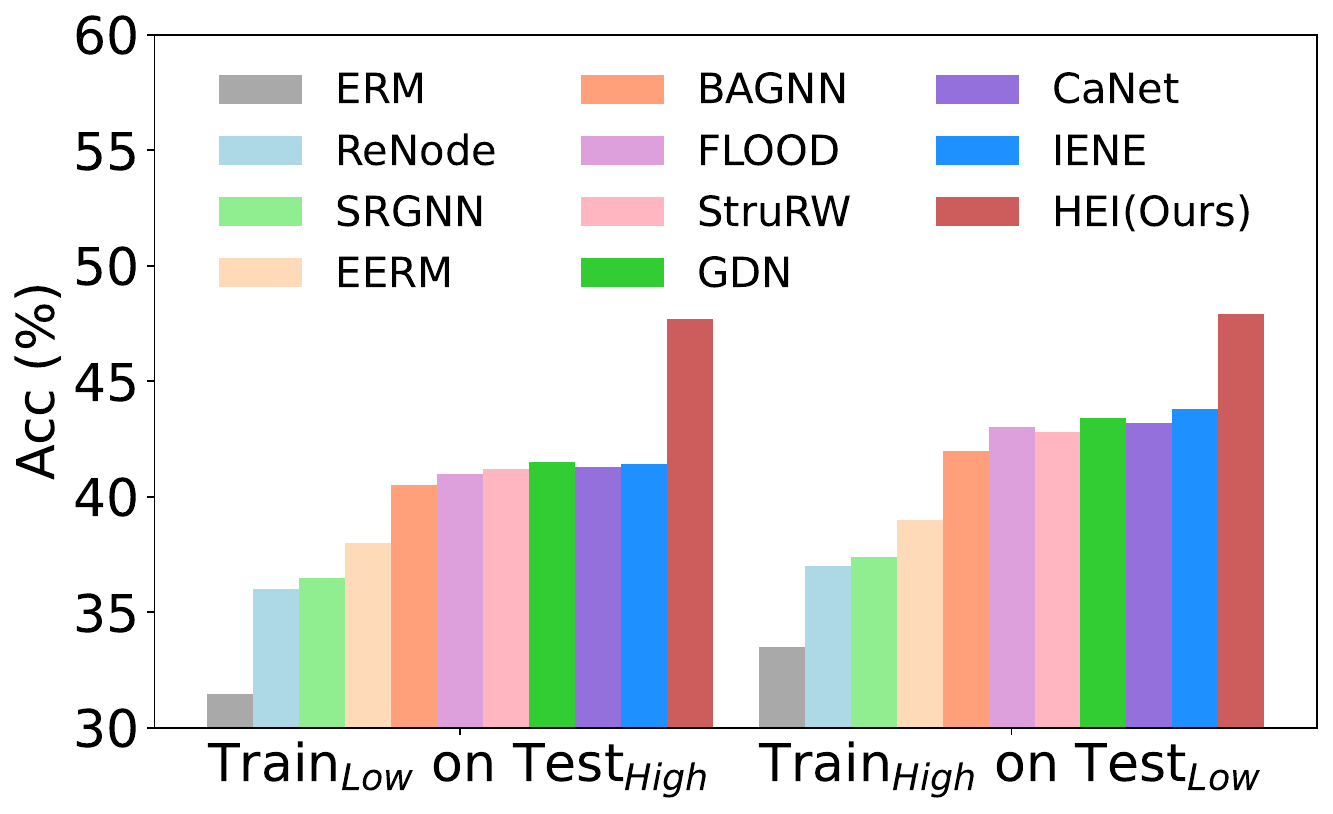}
}
\subfigure[Squirrel-filter]{
\includegraphics[width=0.25\linewidth]{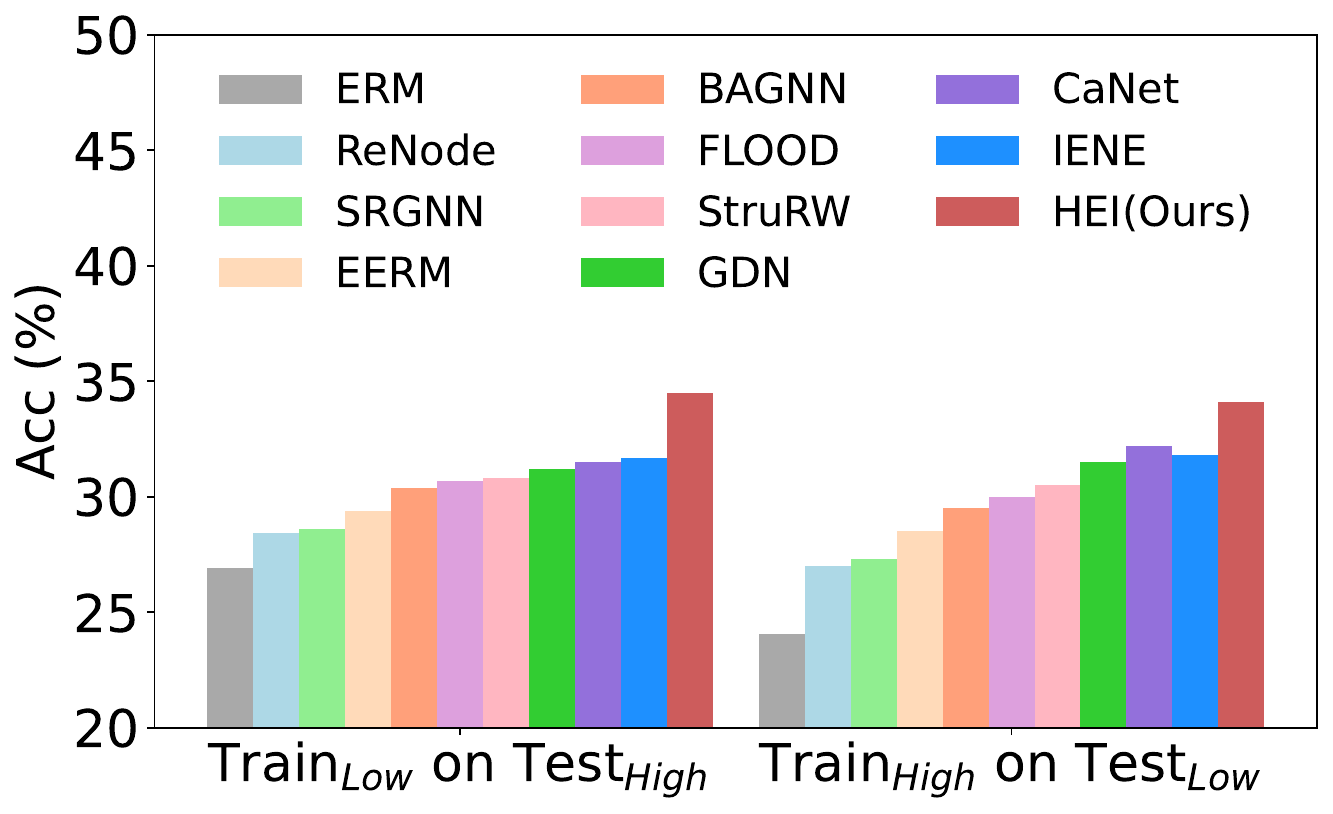}
}
\subfigure[Actor]{
\includegraphics[width=0.25\linewidth]{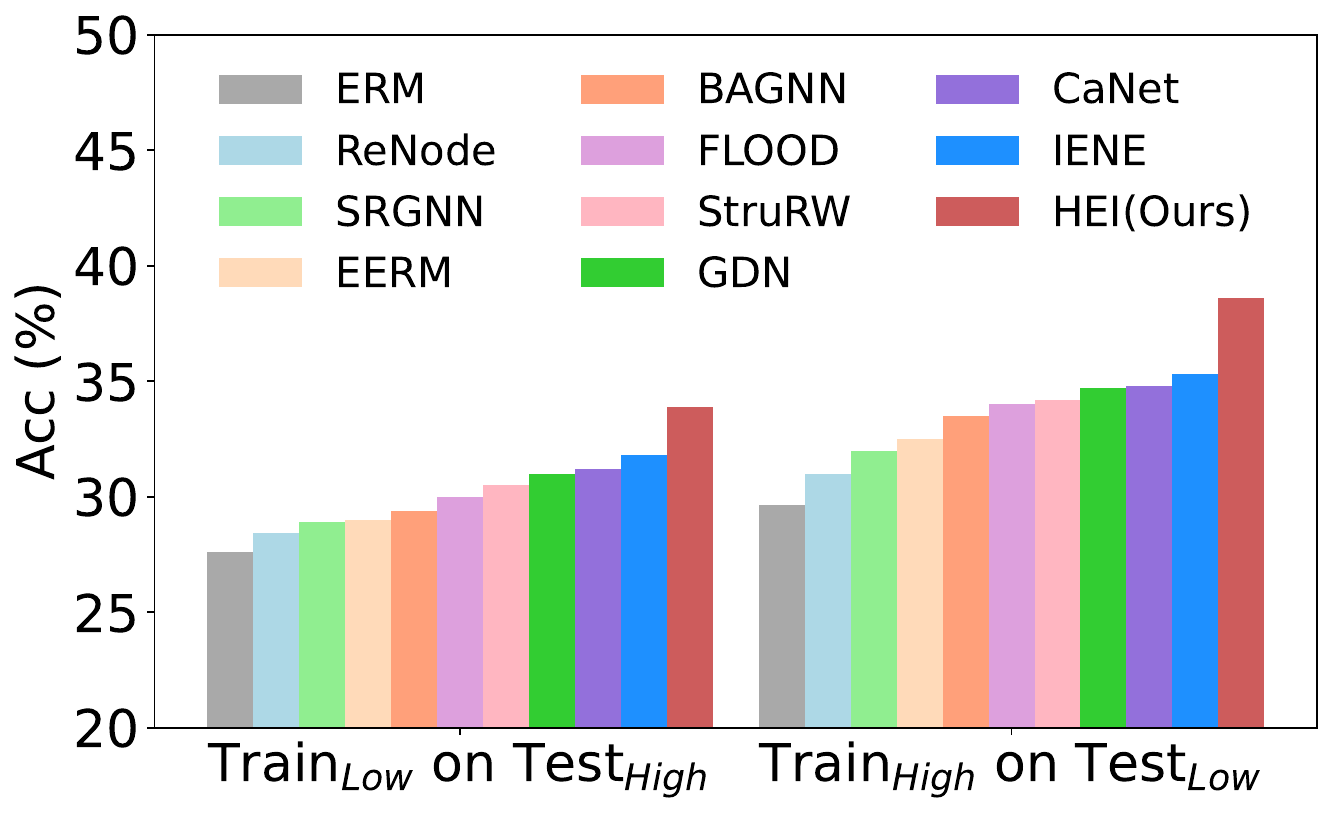}
}
\subfigure[Penn94]{
\includegraphics[width=0.25\linewidth]{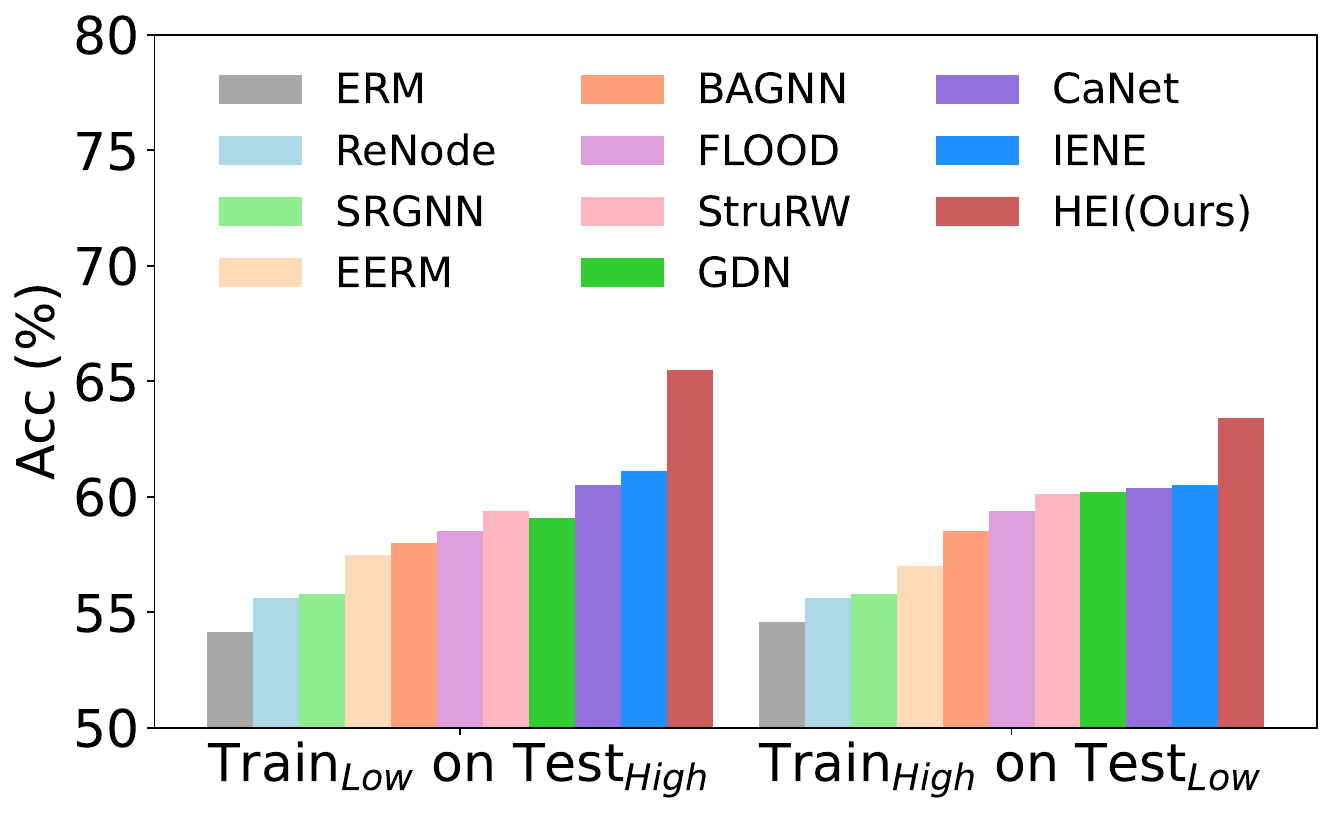}
}
\subfigure[arxiv-year]{
\includegraphics[width=0.25\linewidth]{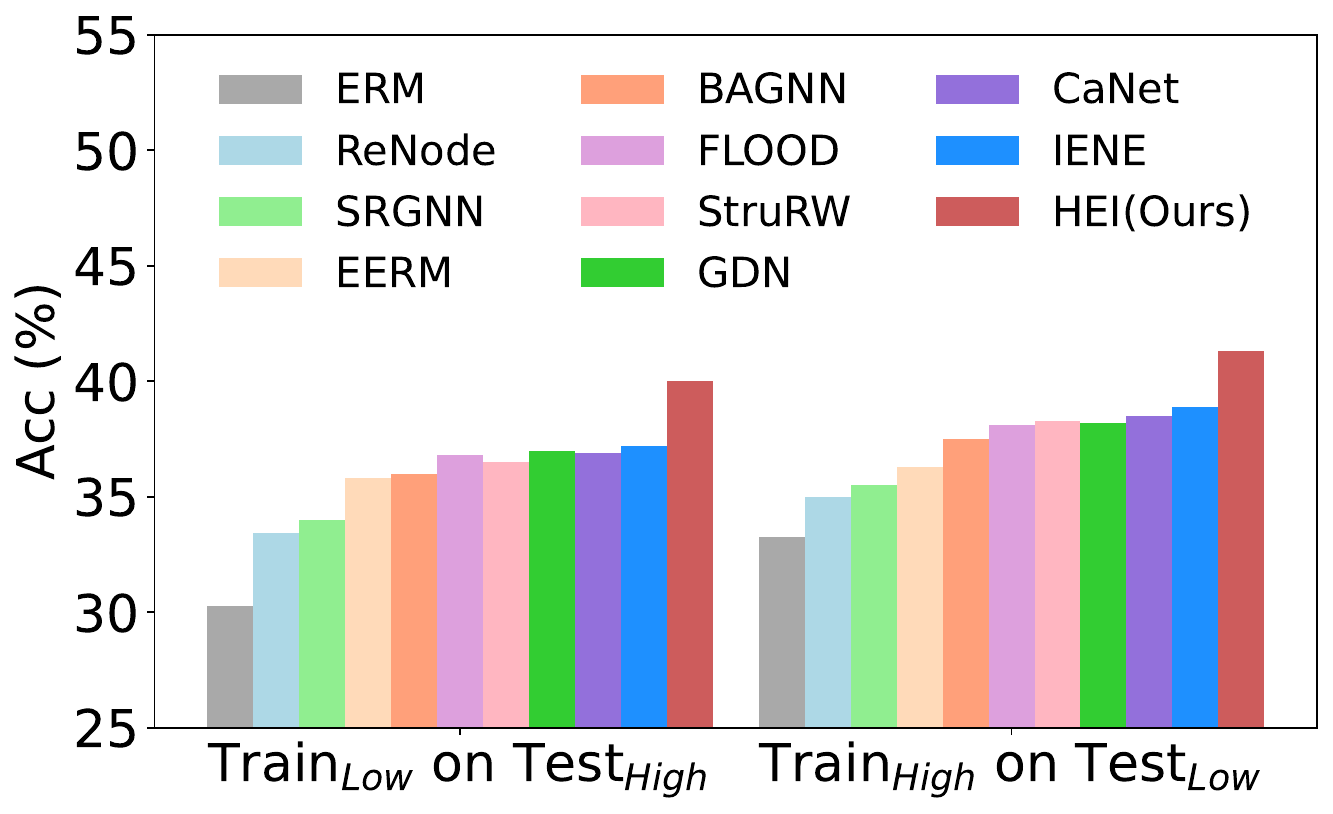}
}
\subfigure[twitch-gamer]{
\includegraphics[width=0.25\linewidth]{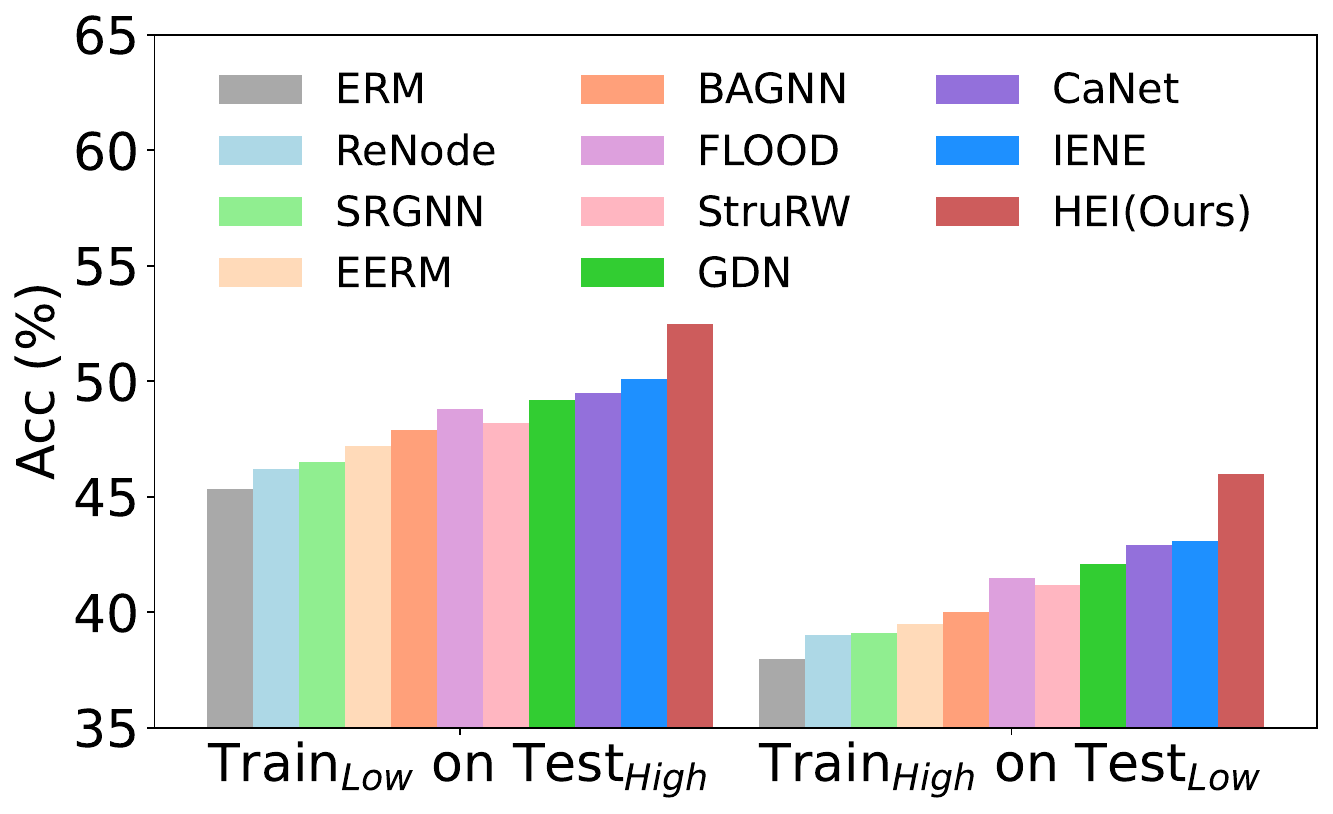}
}
\caption{Comparison experiments under Simulation Settings where exists severe distribution shift between train and test, including $Train_{High}$ on $Test_{Low}$ and $Train_{Low}$ on $Test_{High}$. We adopt the LINKX as the backbone there.}
\Description{Comparison experiments under Simulation Settings where there exists severe distribution shift between train and test, including $Train_{High}$ on $Test_{Low}$ and $Train_{Low}$ on $Test_{High}$. We adopt the LINKX as the backbone there.}
\label{LINKX_large}
\end{center}
\vspace{-0.2cm}
\end{figure*}

\begin{table*}
\setlength{\abovecaptionskip}{0cm}
\setlength{\belowcaptionskip}{0cm}
\captionsetup{font={small,stretch=1.25}, labelfont={bf}}
 \renewcommand{\arraystretch}{1}
    \caption{Comparison experiments on small-scale datasets when we respectively adopt local similarity(Local Sim), post-aggregation similarity (Agg Sim), and SimRank as indicators to estimate nodes' neighbor patterns so as to infer latent environments under Standard settings.  }
    \centering
    \resizebox{0.9\textwidth}{!}{
        \begin{tabular}{ll|l|c|c|c|c|c|c|c|c|c}
            \toprule[1.5pt]
            {}&\textbf{\multirow{2}*{Backbones}} &\textbf{\multirow{2}*{Methods}} &\multicolumn{3}{|c|}{Chamelon-filter} 
            &\multicolumn{3}{|c|}{Squirrel-filter} &\multicolumn{3}{c}{Actor} \\\cline{4-12} 
            \textbf{} & \textbf{} &\textbf{} &Full Test &High Hom Test &Low Hom Test &Full Test  &High Hom Test & Low Hom Test&Full Test  &High Hom Test &Low Hom Test\\ 
            \toprule[1.0pt]
            &\multirow{3}*{LINKX} & HEI (Local Sim)&24.91 ± 3.68 &30.84 ± 3.52  &18.95 ± 3.77 &38.99 ± 1.85 &46.81 ± 1.87 &30.15 ± 1.95 &36.84 ± 1.75&38.99 ± 2.44&34.51 ± 1.88  \\
            &\textbf{}&HEI (Agg Sim)&25.11 ± 3.81 &30.99 ± 3.92  &19.17 ± 3.99 &39.15 ± 1.99 &47.15 ± 1.97 &30.85 ± 2.11 &36.87 ± 1.54&39.11 ± 2.48&34.68 ± 1.74\\
            &\textbf{}&HEI (SimRank) &\textbf{25.78 ± 2.23}	&\textbf{31.35 ± 2.56}	&\textbf{20.21 ± 4.21}& \textbf{40.92 ± 1.31}	&\textbf{48.82 ± 2.88}	&\textbf{31.21 ± 2.79} & \textbf{37.41 ± 1.17}	&\textbf{39.31 ± 1.45}&\textbf{35.42 ± 1.54}\\
            \toprule[1.0pt]
            
            &\multirow{3}*{GloGNN++} & HEI (Local Sim) &27.78 ± 3.51 & 33.25 ± 3.41 &22.51 ± 2.17 &37.55 ± 1.55 &44.11 ± 1.41 &30.11 ± 1.28 &38.57 ± 1.59&41.84 ± 1.44 & 35.41 ± 2.41  \\ 
            &\textbf{}&HEI (Agg Sim) &28.24 ± 3.27 & 33.74 ± 3.11 &22.94 ± 2.27 &37.99 ± 1.25 &44.54 ± 1.28 &30.66 ± 1.51&38.81 ± 1.64&41.95 ± 1.55 & 35.81 ± 2.23\\
            &\textbf{}&HEI (SimRank)  &\textbf{29.31 ± 3.68} & \textbf{34.35 ± 3.52} &\textbf{24.25± 2.71} &\textbf{39.42 ± 1.51} &\textbf{45.19 ± 1.57} &\textbf{31.45 ± 1.68}&\textbf{39.41 ± 1.51} & \textbf{42.25 ± 1.59} & \textbf{36.12 ± 1.85}\\
            \toprule[1.5pt]    
        \end{tabular}
    }
    \label{estimated_small}
\vspace{-0.2cm}
\end{table*}
\noindent \textbf{Handling Distribution Shifts under Simulation Settings where exists severe distribution shifts (RQ2).}
As shown in Figure \ref{LINKX_large},
for each dataset, we mainly report results under severe distribution shifts between training and testing nodes, which include $Train_{High}$ on $Test_{Low}$ and $Train_{Low}$ on $Test_{High}$. We can observe that HEI achieves better performance than other baselines apparently, with up to $5\sim10$ scores on average. In contrast, previous methods with environment augmentation achieved similar improvements. This is because 
all of them succeed in the augmentation-based environmental construction strategy on the ego-graph of train nodes to help the model adapt to diverse distribution, which may be ineffective under the HGSS scenarios, especially when there exists a huge structural pattern gap between train and test nodes. These results can verify the robustness and effectiveness of our proposed HEI.\\

\noindent \textbf{The effect of different similarity metrics as neighbor pattern indicators for HEI (RQ3).} 
As depicted in Table \ref{estimated_small}, we can conclude: (i) Applying any similarity-based metrics can outperform the previous SOTA strategy. This verifies the flexibility and effectiveness of HEI and helps distinguish HEI from previous HGNN works that also utilize the similarity for backbone designs; (ii) Applying SimRank to HEI can acquire consistently better performance than other metrics. This can be explained by previous HGNN backbone designs \cite{luan2022revisiting,chen2023lsgnn,liu2023simga}, which have verified that SimRank has a better ability to distinguish neighbors patterns compared with Local Sim and Agg-Sim, so as to design a better HGNN backbone, SIMGA \cite{liu2023simga}. Moreover, from the perspective of definitions as Eq. \ref{definitions}, the SimRank is specifically designed considering structural information, which is more related to structure-related distribution shifts.  \\

\begin{figure}
\centering
\vspace{-0.2cm}  
\setlength{\abovecaptionskip}{0cm}   
\setlength{\belowcaptionskip}{0cm}   
\subfigure[Chameleon-filter]{
\includegraphics[width=0.3\linewidth]{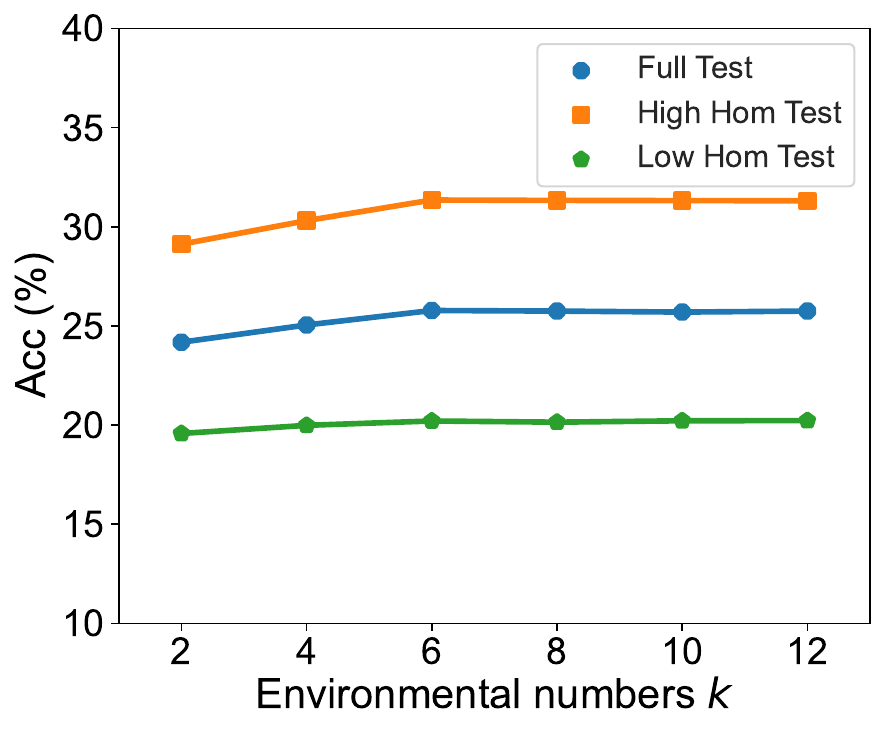}
}
\subfigure[Squirrel-filter]{
\includegraphics[width=0.3\linewidth]{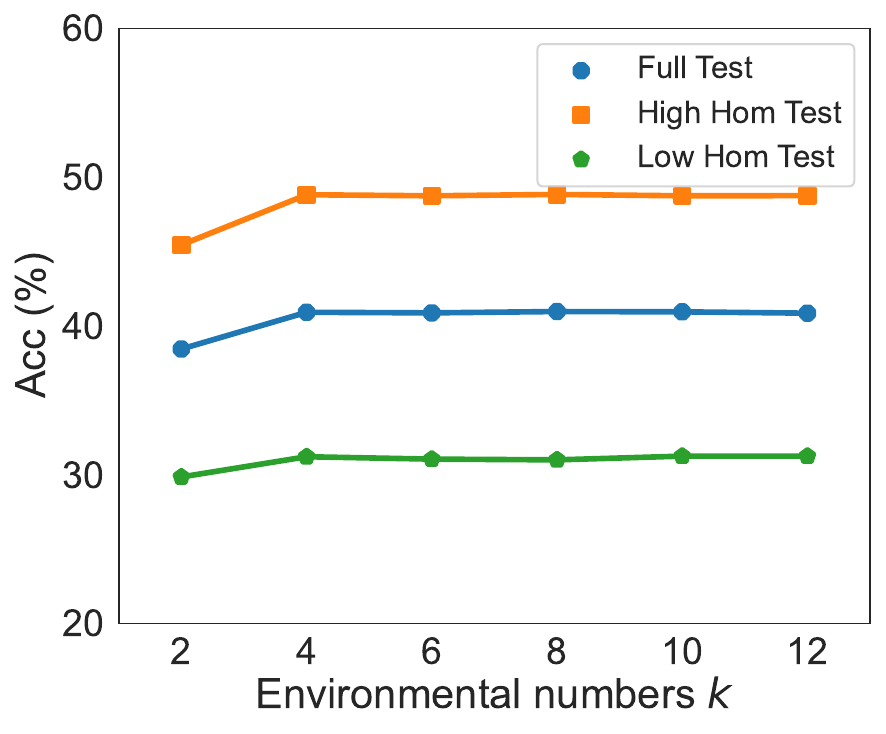}
}
\subfigure[Actor]{
\includegraphics[width=0.3\linewidth]{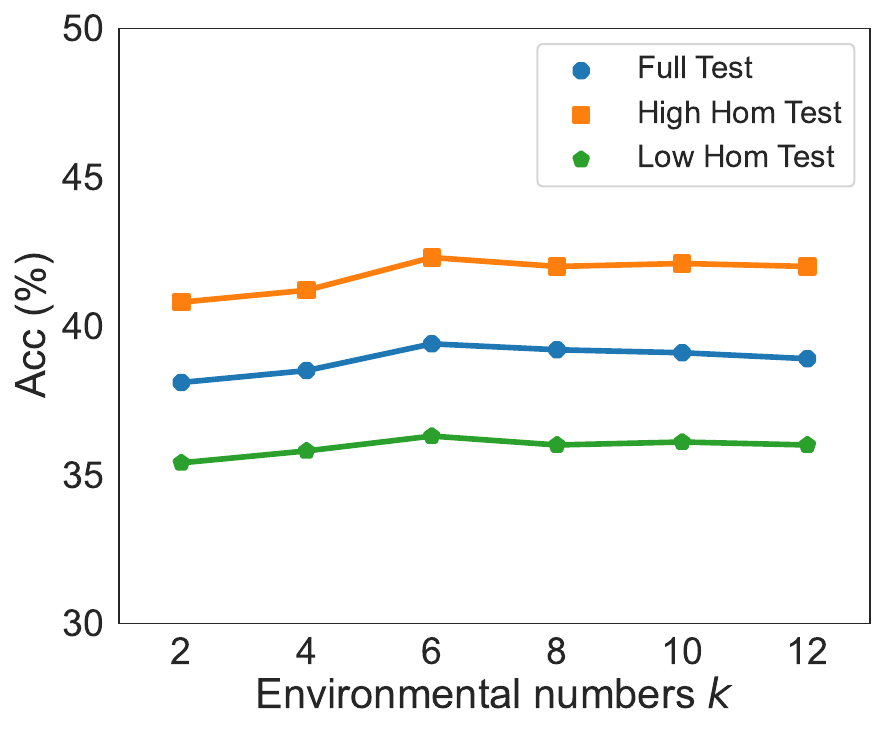}
}
\caption{ Parameter Sensitivity of environmental numbers $k$  under Standard Settings.}
\Description{Parameter Sensitivity of environmental numbers $k$  under Standard Settings.}
\label{env}
\vspace{-0.4cm}
\end{figure}

\noindent \textbf{Sensitivity Analysis of HEI concerning the pre-defined environmental numbers $K$ (RQ4).} 
As shown in Figure \ref{env}, we vary the environment numbers $k$ in Eq. \ref{penalty} within the range of  $\left[2,12\right]$, and keep all other configurations unchanged to explore its impact.  We can observe that  $K$ has a stable effect on the HEI, especially when $k\geq 6$, verifying its effectiveness in addressing HGSS issues.\\



\section{Conclusion}
In this paper, we emphasize an overlooked yet important variety of graph structure distribution shifts that exist on heterophilic graphs. We verify that previous node-level invariant learning solutions with environment augmentation are ineffective due to the irrationality of constructing environments. To mitigate the effect of this distribution shift, we propose HEI, a framework capable of generating invariant node representations by incorporating the estimated neighbor pattern information to infer latent environments without augmentation, which are then used for downstream invariant learning. Experiments on several benchmarks and backbones demonstrate the effectiveness of our method to cope with this graph structure distribution shift. Finally, we hope this study can draw attention to the structural distribution shift of heterophilic graphs.

\begin{acks}
This work was supported by the National Natural Science Foundation of China (62376243, 62441605, 62037001), and the Starry Night Science Fund at Shanghai Institute for Advanced Study (Zhejiang University). 
\end{acks}

\bibliographystyle{ACM-Reference-Format}
\bibliography{sample-base}


\appendix
\newpage
In this appendix, we provide the details omitted in the main text due to the page limit, offering additional experimental results, analyses, proofs, and discussions.

\section{More Details About the Method} \label{appendix_method}

\textbf{Training Process:} As shown by Algorithm \ref{HEI}: Given a heterophilic graph input, we first estimate the neighbor patterns for each node by Eq. \ref{pattern}. Then, based on Eq. \ref{penalty}, we collectively learn the partition of the environment and the invariant representation assisted by the estimated neighbor patterns, to address the HGSS issues.

\begin{figure*}
\begin{center}
\centering
\setlength{\abovecaptionskip}{0cm}   
\setlength{\belowcaptionskip}{0cm}   
\includegraphics[width=0.75\linewidth]{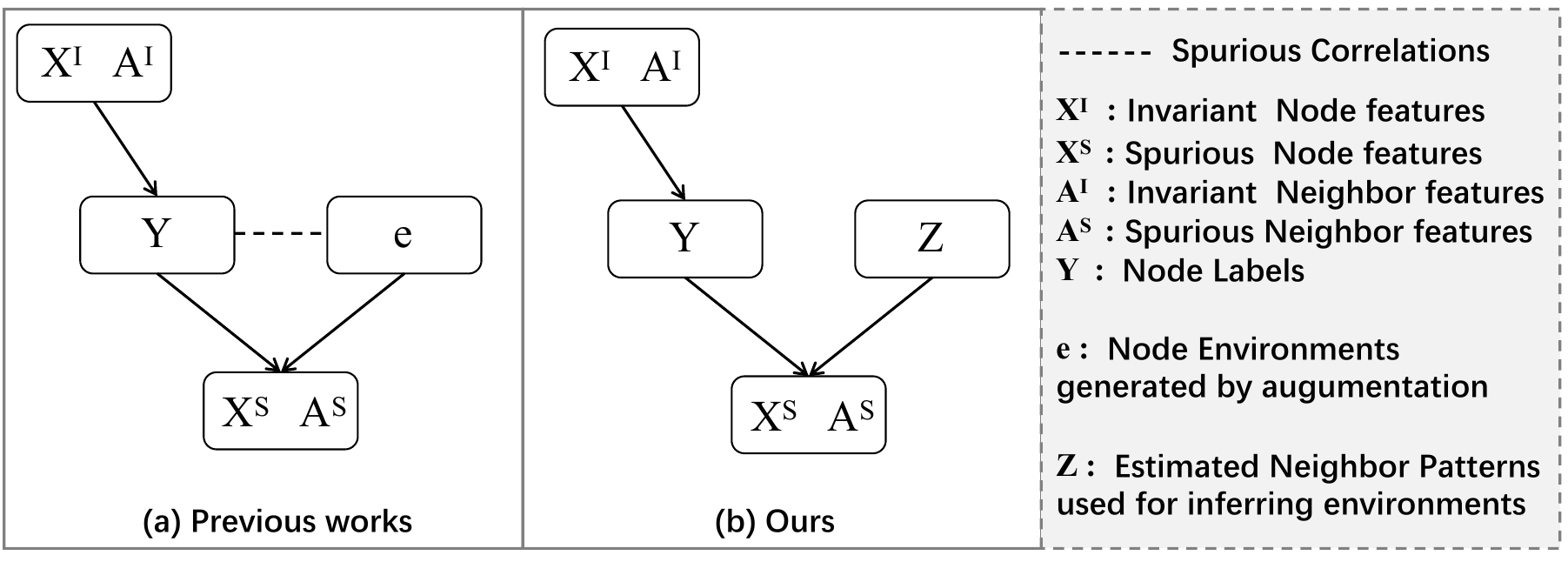}
\caption{Comparisons between our work and previous graph-based invariant learning works from the causal perspective. Notably, the basic HGNN directly aggregates the selected neighbors' full features without further separating like the above two types of invariant learning methods.}
\Description{Comparisons between our work and previous graph-based invariant learning works from the causal perspective.}
\label{causal graph}
\end{center}
\vspace{-0.2cm}
\end{figure*}

\noindent \textbf{Causal Analysis:} To help understand our framework well, we first provide the comparison between our work and previous graph-based invarinat learning works as shown in Figure \ref{causal graph}. Specifically, the definitions of random variables can be defined as follows:
We define $\mathbf G$ as a random variable of the input graph, $\mathbf A$ as a random variable of node's neighbor information, $\mathbf X$ as a random variable of node's features, and $\mathbf Y$ as a random variable of node's label vectors. Both node features $\mathbf X$ and node neighbor information $\mathbf A$ consist of invariant predictive information that determines the label $\mathbf Y$ and the spurious information influenced by latent environments $\mathbf e$. In this case, we can denote $\mathbf X=\left[\mathbf X^{I}, \mathbf X^{S}\right]$ and $ \mathbf A=\left[\mathbf A^{I}, \mathbf A^{S}\right]$. 
Then, we provide a more detailed theoretical analysis of our framework from a casual perspective to identify invariant features.

Casual conditions of $Z$.
Denote the $H(Y|X, A)$ as the expected loss of an optimal classifier over ($X$, $A$, and $Y$), we can clarify the reasonability of utilizing the estimated neighbor patterns as $Z$ auxiliary information to infer environments for invariant prediction based on the following two conditions.

\begin{condition}[Invariance Preserving Condition] 
\label{cond: invariance}
Given invariant feature $(X^{I}$ and $A^{I})$ and any function $\rho(\cdot)$, it holds that 
\begin{align}
H(Y|(X^{I},A^{I}), \rho(Z)) =   H(Y|(X^{I},A^{I})).
\end{align}
\end{condition}

\begin{condition}
[Non-invariance Distinguishing Condition]
\label{cond:non-inv} 
For any feature $ X^{Sk} \in X^S$ or $A^{Sk} \in A^S$ ,there exists a function $\rho(\cdot)$ and a constant $C>0$ satisfy:
\begin{align}
\label{eqn:violation_condition_2}
H(Y|(X^{Sk},A^{Sk})) - {H}(Y|(X^{Sk},A^{Sk}), \rho(Z)) \geq C.
\end{align}
\end{condition}
Condition \ref{cond: invariance} requires that invariant features $X^{I}$ and $A^{I}$ should keep invariant under any environment split obtained by $\rho(Z)$. Otherwise, if there exists a split where an invariant feature becomes non-invariant, then this feature would introduce a positive penalty as shown in Eq. \ref{penalty} to further promote the learning of invariant node representation. Exactly, Condition~\ref{cond: invariance} can be met only if $H(Y|(X^{I},A^{I}), Z) = H(Y|(X^{I},A^{I}))$, which means the auxiliary variable $Z$ should be d-separated by invariant feature $X^{I}$ and $A^{I}$. Exactly, the estimated neighbor pattern just describes the similarity between the node and its neighbors as Eq. \ref{pattern}, while the label $Y$ only has the direct causal relationship with $X^{I}$ and $A^{I}$ from the causal perspective. This means the Condition \ref{cond: invariance} can be met by adopting the estimated neighbor pattern as auxiliary information $Z$ to construct environments.

Condition~\ref{cond:non-inv} reveals that for each spurious feature $X^{S}$ and $A^{S}$, there exists at least one environment split where this feature demonstrates non-invariance within the split environment. If a spurious feature doesn't cause invariance penalties in all environment splits, it can't be distinguished from true invariant features. As shown in Figure \ref {intro}(c2), the results of V-Rex are better than ERM, which means even randomly split environments with seeds can have a positive effect on making spurious features produce effective invariance penalty, further promoting the learning of invariant features. It's more likely to construct comparable or better environments than random seeds under the guidance of the estimated neighbor patterns $Z$. Thus, Condition~\ref{cond:non-inv} can also be guaranteed under our defined heterophilic graph structure distribution shit.

\textbf{Proof of Meeting Condition \ref{cond: invariance}.} We show that for all $\rho(\cdot)$, if $H(Y|(X^{I},A^{I}), Z) = H(Y|(X^{I},A^{I}))$ holds, then there will exist that $H(Y|(X^{I},A^{I}), \rho(Z)) = H(Y|(X^{I},A^{I}))$.
\begin{proof}\renewcommand{\qedsymbol}{}
 On one hand, because $\rho(Z)$ contains less information than $Z$, we have
\[H(Y|(X^{I},A^{I}), \rho(Z)) \geq H(Y|(X^{I},A^{I}), Z) = H(Y|(X^{I},A^{I})). \]
 On the other hand, $(X^{I},A^{I})$ and $\rho(Z)$ contain more information than $(X^{I},A^{I})$, so we can get
  \[H(Y|(X^{I},A^{I}), \rho(Z)) \leq  H(Y|(X^{I},A^{I})). \]
Thus, we conclude $H(Y|(X^{I},A^{I}), \rho(Z)) = H(Y|(X^{I},A^{I}))$.
\end{proof}

\begin{figure}
\begin{minipage}{0.47\textwidth}
\begin{algorithm}[H]
\caption{HEI: Heterophily-Guided Environment inference for Invariant Learning}
\Description{HEI}
\label{HEI}
\begin{algorithmic}[1]
\STATE \textbf{Require:} Graph data $G$ and label $Y$; Environment classifier $\rho$; GNN feature encoder $f_{\Phi}$; GNN classifier $f_{\omega}$; Number of training environments: $K$; a set of Environment-independent GNN classifiers $\{f_{\omega_k}\}_{k=1}^K$;
\STATE Acquire the neighbor pattern estimation $z$ for each node by Eq. \ref{pattern};
\WHILE{Not converged or maximum epochs not reached}
\STATE Divide the nodes into $k$ environments by $\rho(k)(z)$ and obtain corresponding split graphs $\{G_e=k\}_{k=1}^K$;
\FOR{$e=1, \cdots, K$}
\STATE Calculate the GNN's loss on the train nodes belonging to the $k$-th environment, $R_{\rho(k)}(\omega, \Phi)$, via Eq. \ref{erm_enviroments};
\STATE Train an additional environment-independent GNN classifier $f_{\omega_k}$ with the shared GNN feature encoder $f_{\Phi}$ on the train nodes belonging to the $k$-th inferred environment, calculate its loss $R_{\rho(k)}(\omega_k, \Phi)$;
\ENDFOR
\STATE Calculate invariance penalty and the total loss via Eq. \ref{penalty};
\STATE Update $\rho$ via maximizing the invariance penalty;
\STATE Update $f_{\Phi}$, $f_{\omega}$, $f_{\omega}$ via minimizing the total loss;
\ENDWHILE
\end{algorithmic}
\end{algorithm}
\end{minipage}
\end{figure}

\section{More Experimental Results}\label{appendix_experiments}

\noindent \textbf{RQ2: Additional Experiments on Simulation Settings using GloGNN++ as backbone.} We conduct experiments under severe distribution shifts using GloGNN++ as the backbone. As shown in Figure \ref{GloGNN_large}, our proposed method can acquire superior or comparable results than previous methods to handle HGSS, which further verifies the effectiveness and robustness of our design.

\noindent \textbf{RQ3: Effect of different similarity matrices as neighbor pattern indicators for HEI.} We provide large-scale graph experiments as shown in Table \ref{estimated_large} to clarify the details of HEI.

\noindent \textbf{RQ4: Sensitive analysis.} We provide the experimental results about RQ4 there as shown in Figure \ref{env}.

\noindent \textbf{RQ5: Efficiency Studies.}  As shown in Table \ref{efficiency}, referring to \cite{li2022finding}, we provide the time(seconds) to train the model until it converges which keeps the stable accuracy score on the validation set. From the results, we can conclude that the extra time cost can be acceptable compared with the backbone itself.

\noindent \textbf{Experiments on Homophilic Graph Datasets.} Considering the fact that in real-world settings, we can't know whether the input graph is homophilic or heterophilic in advance. Thus, we also provide comparison experiments and discussions for homophilic graphs. As shown in Table \ref{small_homophily}, from the results, we observe that our method can achieve consistent comparable performance to other baselines. But exactly, the improvements by these methods are all minor compared with the results of ERM. That's because the homophilic graph is not related to our settings. After all, homophilic graph datasets mean the neighbor pattern distribution between the train and test are nearly the same, which is not suitable to clarify our defined distribution shifts. The performance gap between the low home test and the high home test can support our analysis.


\begin{table}[ht]
\setlength{\abovecaptionskip}{0cm}
\setlength{\belowcaptionskip}{0cm}
    \captionsetup{font={small,stretch=1.25}, labelfont={bf}}
    \renewcommand{\arraystretch}{1.2}
    \caption{Efficiency studies of HEI on GloGNN++.} 
    \centering
    \setlength{\abovecaptionskip}{0cm}   
    \setlength{\belowcaptionskip}{0cm}   
    \resizebox{0.4\textwidth}{!}{
        \begin{tabular}{l|c|c|c}
            \toprule[1.5pt]
            \textbf{Methods} & \textbf{Penn94} & \textbf{arxiv-year} & \textbf{twitch-gamer} \\
            \toprule[1.0pt]
            ERM & 22.3 & 7.2 & 40.5 \\  
            Renode & 23.5 & 8.5 & 41.2 \\  
            SRGNN & 24.9 & 9.1 & 41.0 \\  
            EERM & 24.7 & 8.8 & 41.5 \\  
            BAGNN & 24.8 & 9.1 & 42.1 \\  
            FLOOD & 24.5 & 8.8 & 41.8 \\  
            StruRW & 23.8 & 9.8 & 41.5 \\ 
            GDN & 24.7 & 9.6 & 42.2 \\
            CaNet & 25.9 & 10.8 & 42.2 \\ 
            IENE & 25.4 & 10.6 & 42.8 \\ 
            HEI(Ours) & 26.8 & 11.5 & 44.9 \\ 
            \toprule[1.5pt] 
        \end{tabular}
    }
    \label{efficiency}
\end{table}

\begin{table*}
\setlength{\abovecaptionskip}{0cm}
\setlength{\belowcaptionskip}{0cm}
\captionsetup{font={small,stretch=1.25}, labelfont={bf}}
 \renewcommand{\arraystretch}{1}
    \caption{Performance comparison on homophilic graph datasets under Standard Settings. The reported scores denote the classification accuracy (\%) and error bar (±) over 10 trials.  We highlight the best score on each dataset in bold and the second score with underline. }
    \centering
    \resizebox{0.7\textwidth}{!}{
        \begin{tabular}{ll|c|c|c|c|c|c|c|c|c|c}
            \toprule[1.5pt]
            {}&\textbf{\multirow{2}*{Backbones}} &\textbf{\multirow{2}*{Methods}} &\multicolumn{3}{|c|}{CiteSeer} 
            &\multicolumn{3}{|c|}{PubMed} &\multicolumn{3}{c}{Cora} \\\cline{4-12} 
            \textbf{} & \textbf{} &\textbf{} &Full Test  &High Hom Test &Low Hom Test &Full Test  &High Hom Test & Low Hom Test&Full Test  &High Hom Test &Low Hom Test\\ 
            \toprule[1.0pt]
            &\multirow{9}*{LINKX} & ERM   &73.19 ± 0.99 &73.89 ± 1.51   &72.79 ± 1.47 &87.86 ± 0.77	&88.49 ± 1.37	&87.19 ± 1.51&84.64 ± 1.13	&85.13 ± 1.83	&83.85 ± 1.99 \\
            &\textbf{} &{ReNode}&73.25 ± 0.89	&74.00 ± 1.21	&72.89 ± 1.87&87.91 ± 0.72	&88.58 ± 1.42	&87.21 ± 1.71&84.70 ± 1.23	&85.28 ± 1.99	&84.24 ± 2.13\\
            &\textbf{} &{SRGNN}&73.27 ± 0.99	&74.03 ± 1.11	&72.85 ± 1.87&87.96 ± 0.81	&88.68 ± 1.72	&87.31 ± 1.51&84.71 ± 1.25	&85.22 ± 1.87	&84.34 ± 2.03\\
            &\textbf{}&{StruRW-Mixup} &73.29 ± 0.91	&73.93 ± 1.25&72.99 ± 1.91&88.12 ± 0.51	&88.71 ± 1.44	&87.58 ± 1.59&84.67 ± 1.54	&85.33 ± 1.91	&84.34 ± 2.43\\
            &\textbf{}&{EERM}&73.17 ± 0.79	&73.81 ± 1.45	&73.09 ± 1.65&87.96 ± 0.84	&88.59 ± 1.32	&87.29 ± 1.63&84.62 ± 1.37	&85.14 ± 1.63	&83.87 ± 2.03\\
            &\textbf{}&{BAGNN}&73.33 ± 0.88	&73.99 ± 1.61	&73.19 ± 1.63&88.01 ± 0.94	&88.78 ± 1.57	&87.69 ± 1.39&84.60 ± 1.28	&85.24 ± 1.83	&83.84 ± 2.41\\
                &\textbf{}&{FLOOD}&73.34 ± 0.91	&73.95 ± 1.55	&73.22 ± 1.67&88.05 ± 0.95&88.84 ± 1.62	&87.81 ± 1.59&84.72 ± 1.41	&85.35 ± 1.63	&83.99 ± 2.51\\
                &\textbf{}&{CaNet}&73.38 ± 0.95	&74.08 ± 1.54	&73.31 ± 1.55&88.11 ± 0.98&88.89 ± 1.67	&87.91 ± 1.54&84.81 ± 1.31	&85.39 ± 1.57	&84.11 ± 2.58\\
                    &\textbf{}&{IENE}&\underline{73.43 ± 0.97}	&\underline{74.15 ± 1.58}	&\underline{73.32 ± 1.87} &88.12 ± 0.94&88.90 ± 1.65	&87.90 ± 1.66&\underline{84.92 ± 1.45}	&\underline{85.41 ± 1.68}	&\underline{84.45 ± 2.81}\\
                &\textbf{}&{GDN}&73.31 ± 0.81	&73.99 ± 1.61	&73.21 ± 1.68&\underline{88.14 ± 0.94}	&\underline{88.91 ± 1.64}	&\underline{87.92 ± 1.41}&84.64 ± 1.33	&85.27 ± 1.69	&83.91 ± 2.78\\                
            &\textbf{}&\textbf{HEI(Ours)}&\textbf{73.51 ± 0.81}	&\textbf{74.18 ± 1.25}	&\textbf{73.42 ± 1.85}& \textbf{88.50 ± 0.97}	&\textbf{89.01 ± 1.24}	&\textbf{87.99 ± 1.92} & \textbf{85.17 ± 1.53}	&\textbf{85.44 ± 1.83}&\textbf{84.84 ± 1.97}\\
            \toprule[1.0pt]
            
            &\multirow{9}*{GloGNN++}&ERM  &77.22 ± 1.78 & 78.15 ± 2.55 & 76.79 ± 2.54 &89.24 ± 0.39 & 90.62 ± 0.99 & 88.75 ± 1.28 &88.33 ± 1.09 & 90.06 ± 1.52 & 87.37 ± 1.64\\ 
            &\textbf{}&{ReNode}&77.31 ± 1.69 & 78.27 ± 2.48 & 76.90 ± 2.39 &89.25 ± 0.35 & 90.64 ± 0.87 & 88.79 ± 1.24 &88.39 ± 1.21 & 90.11 ± 1.49 & 87.45 ± 1.57\\
            &\textbf{} &{SRGNN}&77.33 ± 1.65 & 78.24 ± 2.75 & 76.91 ± 2.77 &89.33 ± 0.51 & 90.81 ± 1.21 & 88.99 ± 1.57 &88.53 ± 1.09 & 90.46 ± 1.53 & 87.58 ± 1.54\\
            &\textbf{}&{StruRW-Mixup} &77.35 ± 1.57 & 78.27 ± 2.11 & 76.90 ± 2.42 &89.48 ± 0.44 & 90.81 ± 0.97 & 89.17 ± 1.33 &88.39 ± 1.44 & 90.15 ± 1.69 & 87.85 ± 1.77\\
            &\textbf{}&{EERM}&77.35 ± 1.81 & 78.27 ± 2.45 & 76.89 ± 2.81 &89.34 ± 0.39 & 90.82 ± 1.09 & 88.95 ± 1.38 &88.39 ± 1.21 & 90.36 ± 1.42 & 87.47 ± 1.74\\
            &\textbf{}&{BAGNN}&77.42 ± 1.81 & 78.35 ± 2.81 & 76.89 ± 2.42 &89.37 ± 0.45 & 90.87 ± 1.29 & 88.99 ± 1.58 &88.49 ± 1.31 & 90.39 ± 1.47 & 87.81 ± 1.64\\
            &\textbf{}&{FLOOD}&77.43 ± 1.79 & 78.39 ± 2.51 & 76.95 ± 2.37 &89.41 ± 0.51 &90.91 ± 1.29 & 89.08 ± 1.58 &88.51 ± 1.27 & 90.40 ± 1.51 & 87.88 ± 1.69\\

            &\textbf{}&{CaNet}&77.42 ± 1.81 & 78.32 ± 2.54 & 76.75 ± 2.87 &89.51 ± 0.58 &90.97 ± 1.29 & 89.19 ± 1.61 &88.54 ± 1.33 & 90.51 ± 1.61 & 88.11 ± 1.58\\
            
            &\textbf{}&{IENE}&77.43 ± 1.79 & 78.39 ± 2.51 & 76.96 ± 2.58 &\underline{89.52 ± 0.57} &\underline{91.00 ± 1.44} & \underline{89.22 ± 1.62} &\underline{88.78 ± 1.37} & \underline{90.79 ± 1.47} & \underline{88.28 ± 1.49} \\ 
            
            &\textbf{}&{GDN} &\underline{77.44 ± 1.51} & \underline{78.65 ± 3.75} & \underline{76.97 ± 2.53} &89.39 ± 0.51 & 90.97 ± 1.01 & 88.99 ± 1.21 &88.29 ± 1.34 & 90.44 ± 1.91 & 87.89 ± 1.22\\
            &\textbf{}&\textbf{HEI(Ours)}
            &\textbf{77.85 ± 1.89} & \textbf{79.11 ± 2.59} & \textbf{77.30 ± 2.85} & \textbf{89.99 ± 0.39} & \textbf{91.52 ± 0.99} & \textbf{89.48 ± 1.33} &\textbf{88.93 ± 1.19} & \textbf{90.97 ± 1.39} & \textbf{88.47 ± 1.74}\\
            \toprule[1.5pt]    
        \end{tabular}
    }
    \label{small_homophily}
\end{table*}

\begin{table*}
\setlength{\abovecaptionskip}{0cm}
\setlength{\belowcaptionskip}{0cm}
\captionsetup{font={small,stretch=1.25}, labelfont={bf}}
 \renewcommand{\arraystretch}{1}
    \caption{Comparison experiments on large-scale datasets when we respectively adopt local similarity(Local Sim), post-aggregation similarity (Agg Sim), and SimRank as indicators to estimate nodes' neighbor patterns so as to infer latent environments under Standard settings.  }
    \centering
    \resizebox{0.8\textwidth}{!}{
        \begin{tabular}{ll|l|c|c|c|c|c|c|c|c|c}
            \toprule[1.5pt]
            {}&\textbf{\multirow{2}*{Backbones}} &\textbf{\multirow{2}*{Methods}} &\multicolumn{3}{|c|}{Penn94} 
            &\multicolumn{3}{|c|}{arxiv-year} &\multicolumn{3}{c}{twitch-gamer} \\\cline{4-12} 
            \textbf{} & \textbf{} &\textbf{} &Full Test &High Hom Test &Low Hom Test &Full Test  &High Hom Test & Low Hom Test&Full Test  &High Hom Test &Low Hom Test\\ 
            \toprule[1.0pt]
            &\multirow{3}*{LINKX} & HEI (Local Sim)  &85.12 ± 0.21	&88.28 ± 0.33 &82.15 ± 0.59	&54.41 ± 0.21	&64.23 ± 0.47	&48.29 ± 0.22	&66.18 ± 0.12	&83.75 ± 0.34	&48.12 ± 0.47 \\
            &\textbf{}&HEI (Agg Sim)&85.21 ± 0.17	&88.29 ± 0.38	&82.22 ± 0.54	&54.45 ± 0.23	&64.33 ± 0.49	&48.33 ± 0.32 &66.21 ± 0.15	&83..85 ± 0.39	&48.45 ± 0.57\\
            &\textbf{}&HEI (SimRank)&\textbf{86.22 ± 0.28}	&\textbf{89.24 ± 0.28}&\textbf{83.22 ± 0.59} &\textbf{56.05 ± 0.22}&	\textbf{66.53 ± 0.41}	&\textbf{49.33 ± 0.32} &\textbf{66.79 ± 0.14} & \textbf{85.33 ± 0.25} &\textbf{49.21 ± 0.57}\\
            \toprule[1.0pt]
            
            &\multirow{3}*{GloGNN++}& HEI (Local Sim)     &86.08 ± 0.24	&89.70 ± 0.64	&82.18 ± 0.37 &54.42 ± 0.24 &64.48 ± 1.54	&48.55 ± 0.64 &66.30 ± 0.18 &83.21 ± 0.68&49.00 ± 0.67\\ 
            
            &\textbf{}&HEI (Agg Sim)&86.15 ± 0.25	&89.70 ± 0.69	&82.43 ± 0.38 &54.44 ± 0.25 &64.51 ± 1.54	&48.69 ± 0.81 
            &66.34 ± 0.21 &83.19 ± 0.78 &49.14 ± 0.57\\
            
            &\textbf{}&HEI (SimRank) &\textbf{87.18 ± 0.28}	&\textbf{89.99 ± 0.65}	&\textbf{83.59 ± 0.39} &\textbf{55.71 ± 0.24}&\textbf{66.29 ± 1.14}	&\textbf{49.52 ± 0.75} 
            &\textbf{66.99 ± 0.17} &\textbf{84.37 ± 0.68}&\textbf{50.40 ± 0.52}\\
            \toprule[1.5pt]    
        \end{tabular}
    }
    \label{estimated_large}
\end{table*}

\begin{figure}
\begin{center}
\setlength{\abovecaptionskip}{0cm}   
\setlength{\belowcaptionskip}{0cm}   
\subfigure[Chameleon]{
\includegraphics[width=0.27\linewidth]{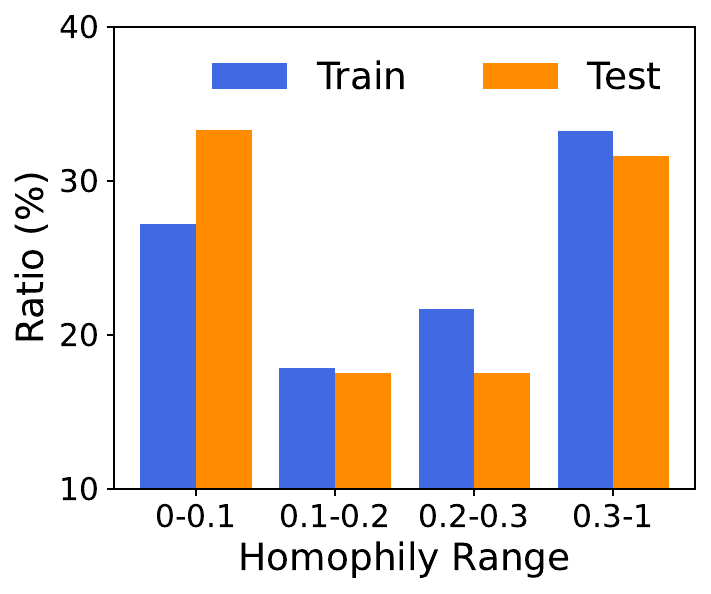}
}
\subfigure[Squirrel]{
\includegraphics[width=0.27\linewidth]{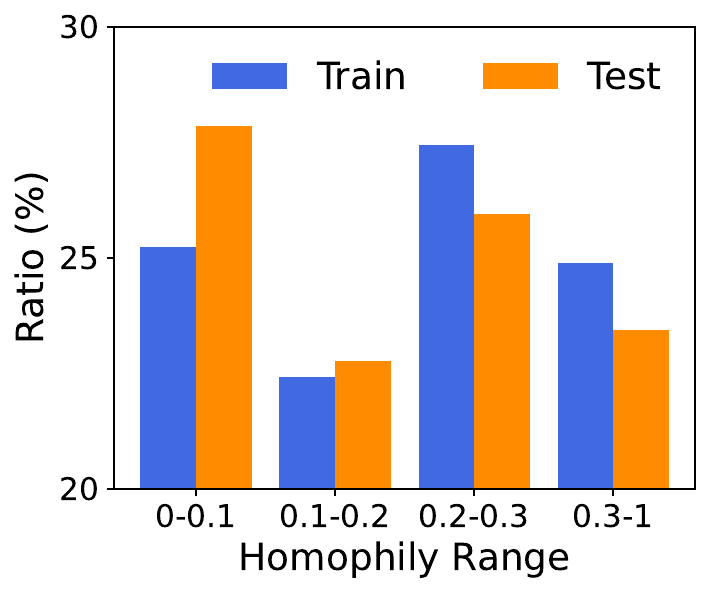}
}
\subfigure[Actor]{
\includegraphics[width=0.27\linewidth]{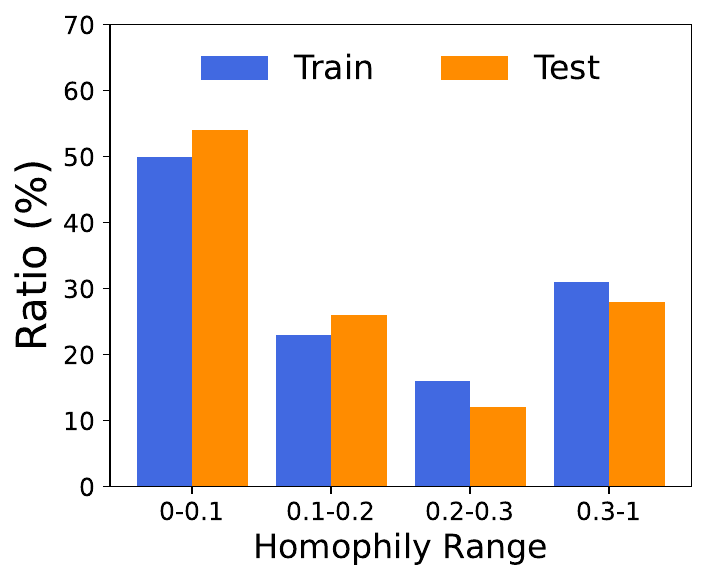}
}
\subfigure[Penn94l]{
\includegraphics[width=0.27\linewidth]{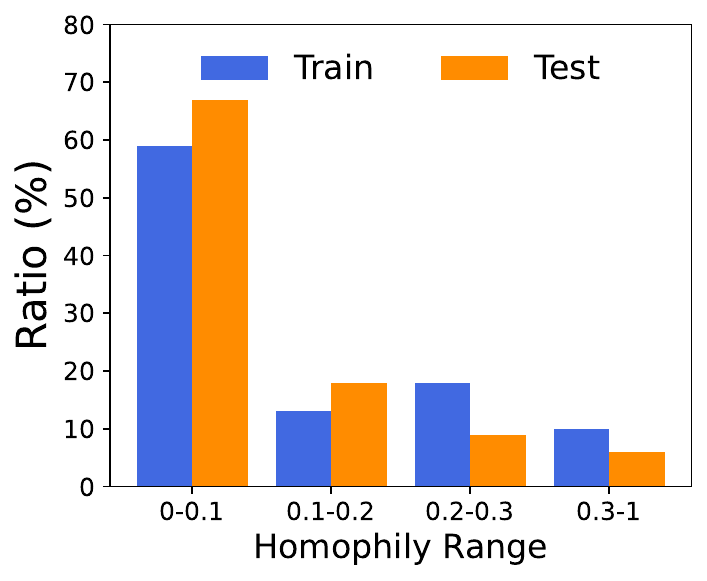}
}
\subfigure[arxiv-year]{
\includegraphics[width=0.27\linewidth]{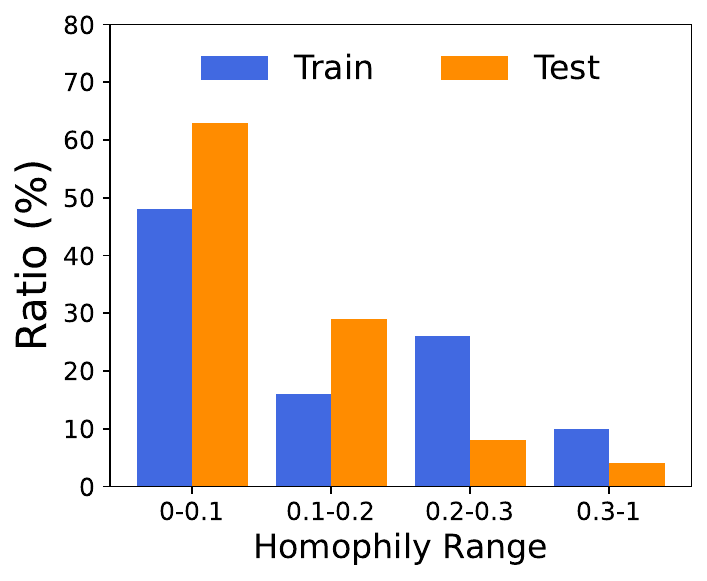}
}
\subfigure[twitch-gamer]{
\includegraphics[width=0.27\linewidth]{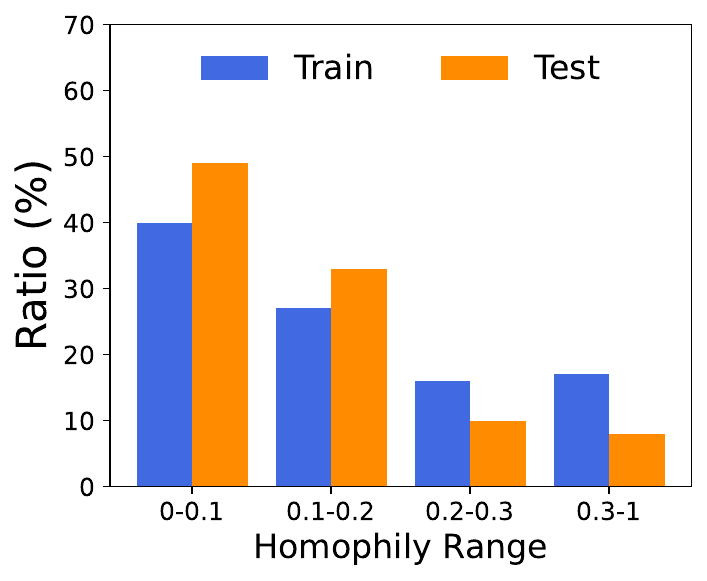}
}
\caption{Statistic of homophily ratio for train and test nodes following previous dataset splits. The nodes are categorized into four groups according to the node-level homophily. Compared with test nodes, the train nodes are more prone to be categorized into groups with high homophily. In other words, in the range with high homophily(from 0.2 to 0.3 and from 0.3 to 1), the sub-train ratio in all train nodes is higher than the sub-test in all test nodes. But in the range with low homophily, there exists a contrary phenomenon.}
\label{intro_hom}
\Description{Statistic of homophily ratio for train and test nodes following previous dataset splits.}
\end{center}
\vspace{-0.4cm}
\end{figure}

\begin{figure*}
\setlength{\abovecaptionskip}{0cm}   
\setlength{\belowcaptionskip}{0cm}   
\begin{center}
\vspace{-2mm}
\subfigure[Chameleon-filter]{
\includegraphics[width=0.25\linewidth]{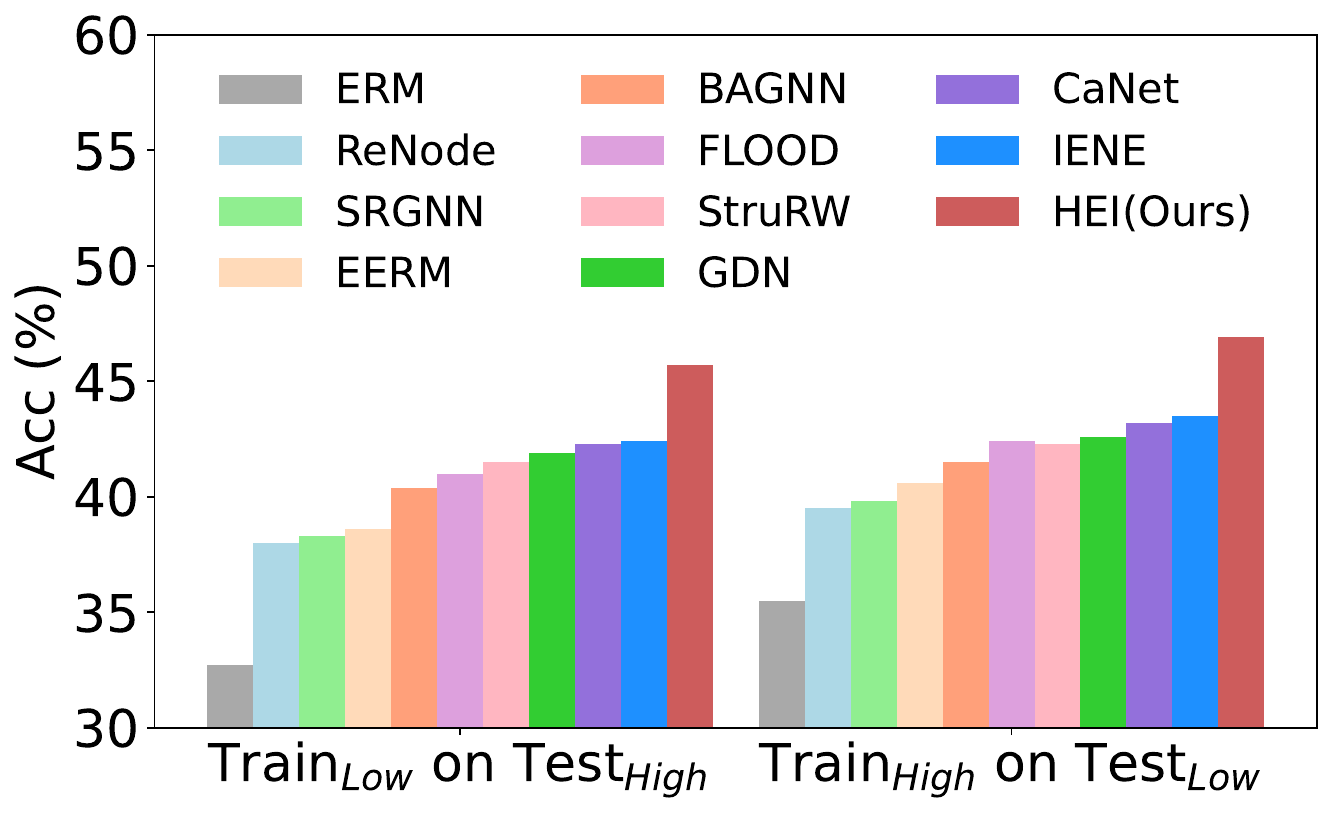}
}
\vspace{-2mm}
\hspace{2mm} 
\subfigure[Squirrel-filter]{
\includegraphics[width=0.25\linewidth]{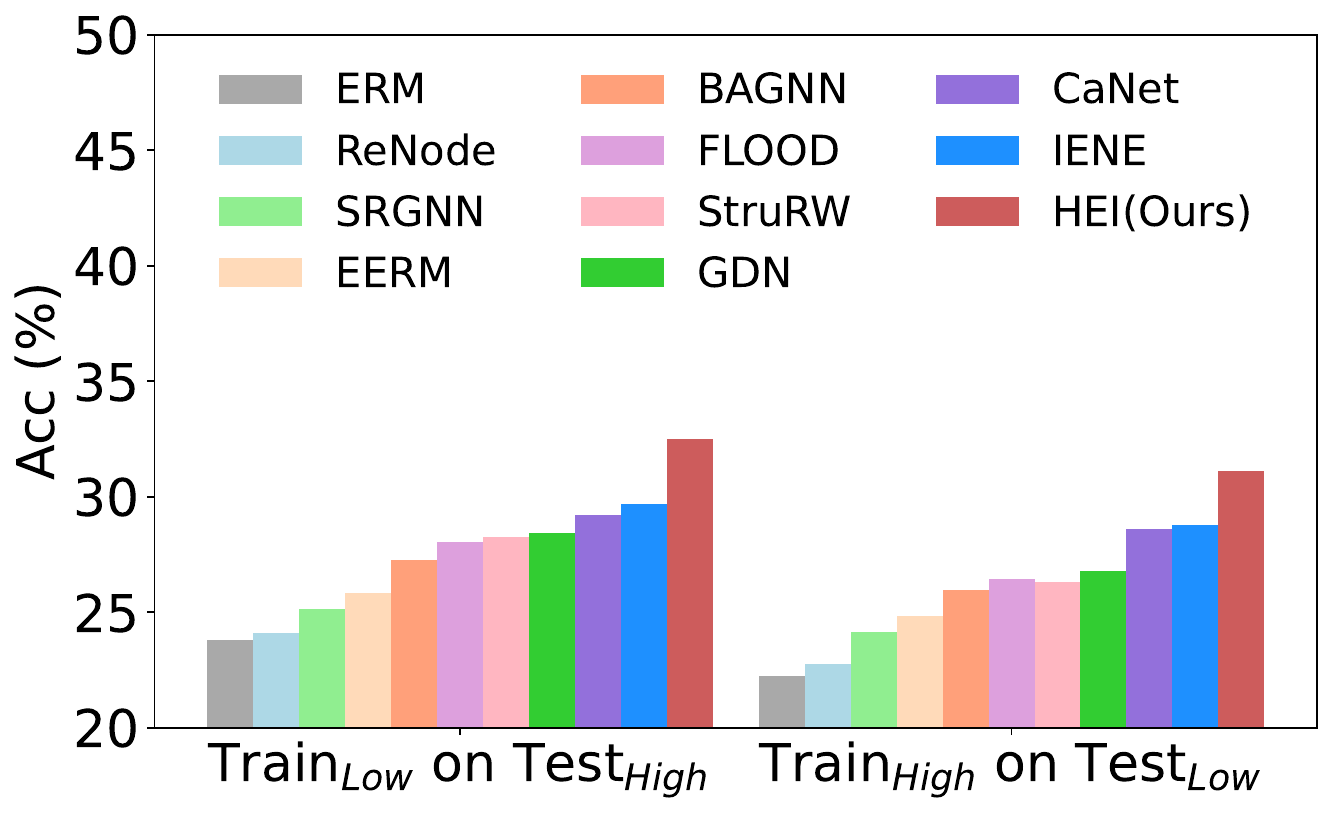}
}
\vspace{-2mm}
\hspace{2mm}
\subfigure[Actor]{
\includegraphics[width=0.25\linewidth]{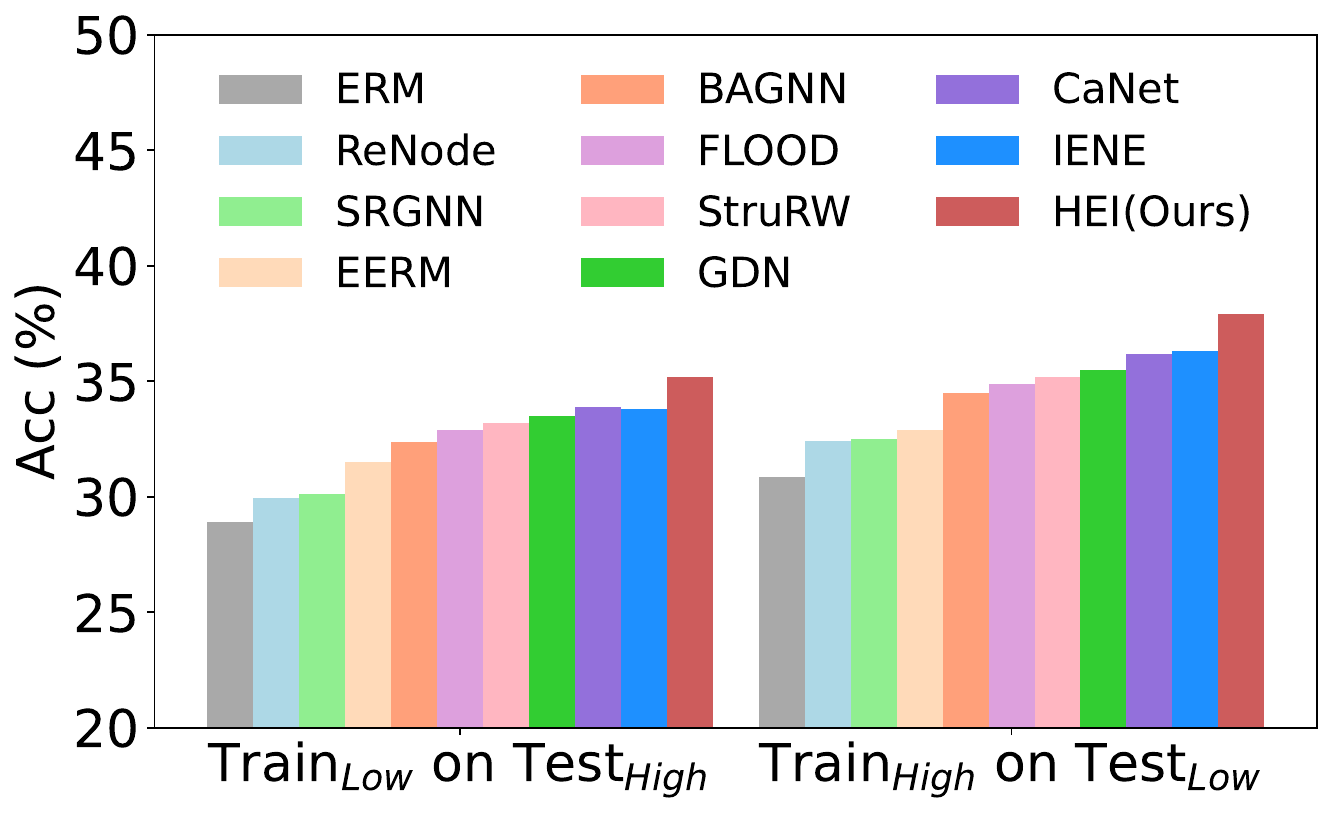}
}
\hspace{2mm}
\subfigure[Penn94]{
\includegraphics[width=0.25\linewidth]{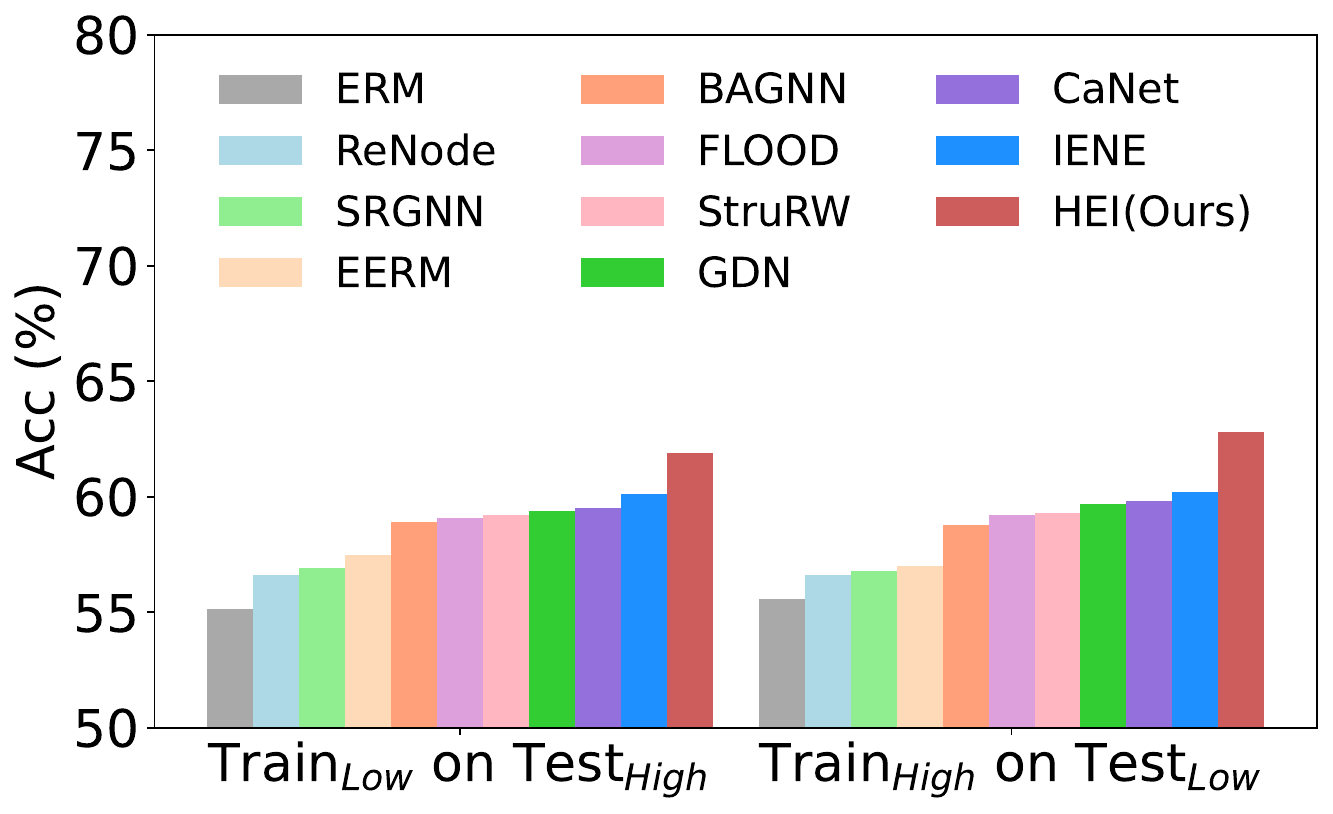}
}
\hspace{2mm}
\subfigure[arxiv-year]{
\includegraphics[width=0.25\linewidth]{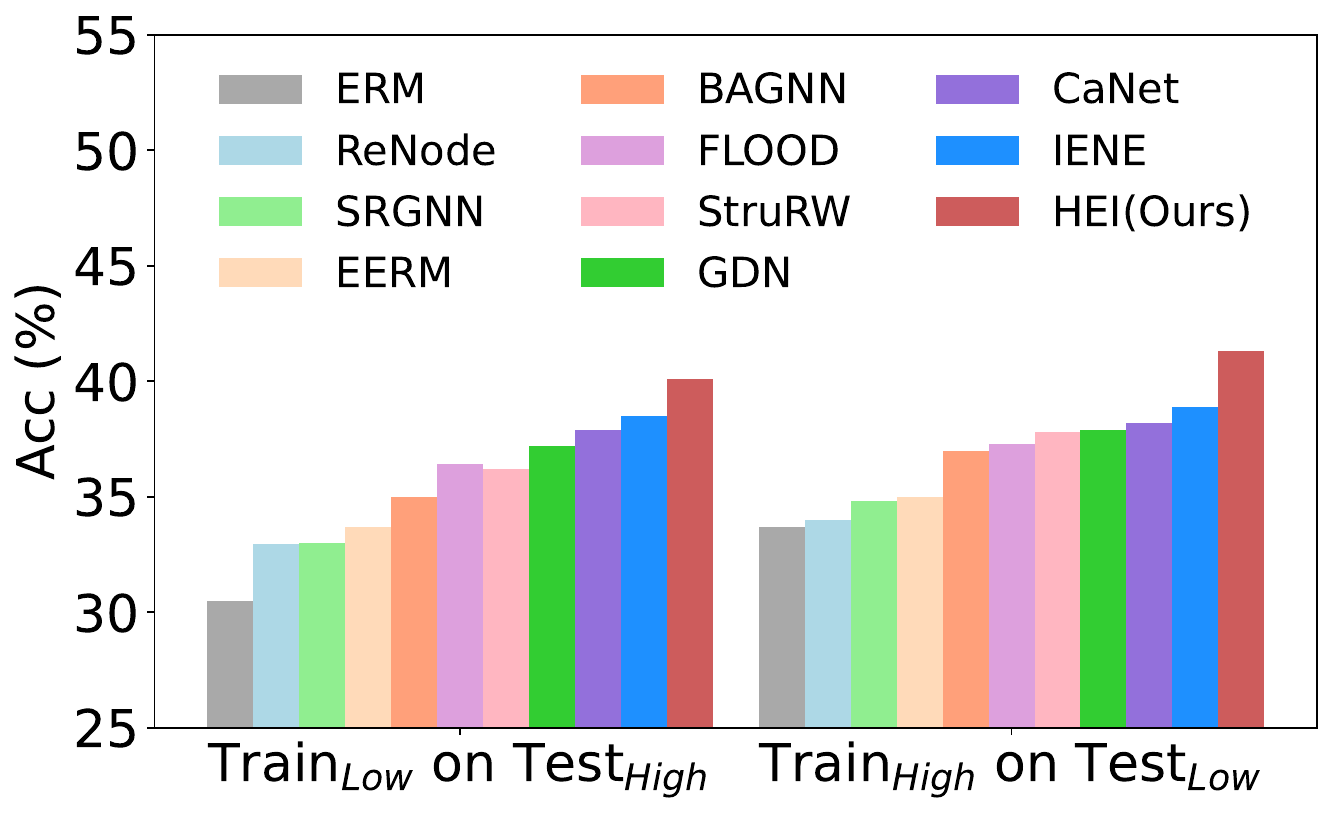}
}
\hspace{2mm}
\subfigure[twitch-gamer]{
\includegraphics[width=0.25\linewidth]{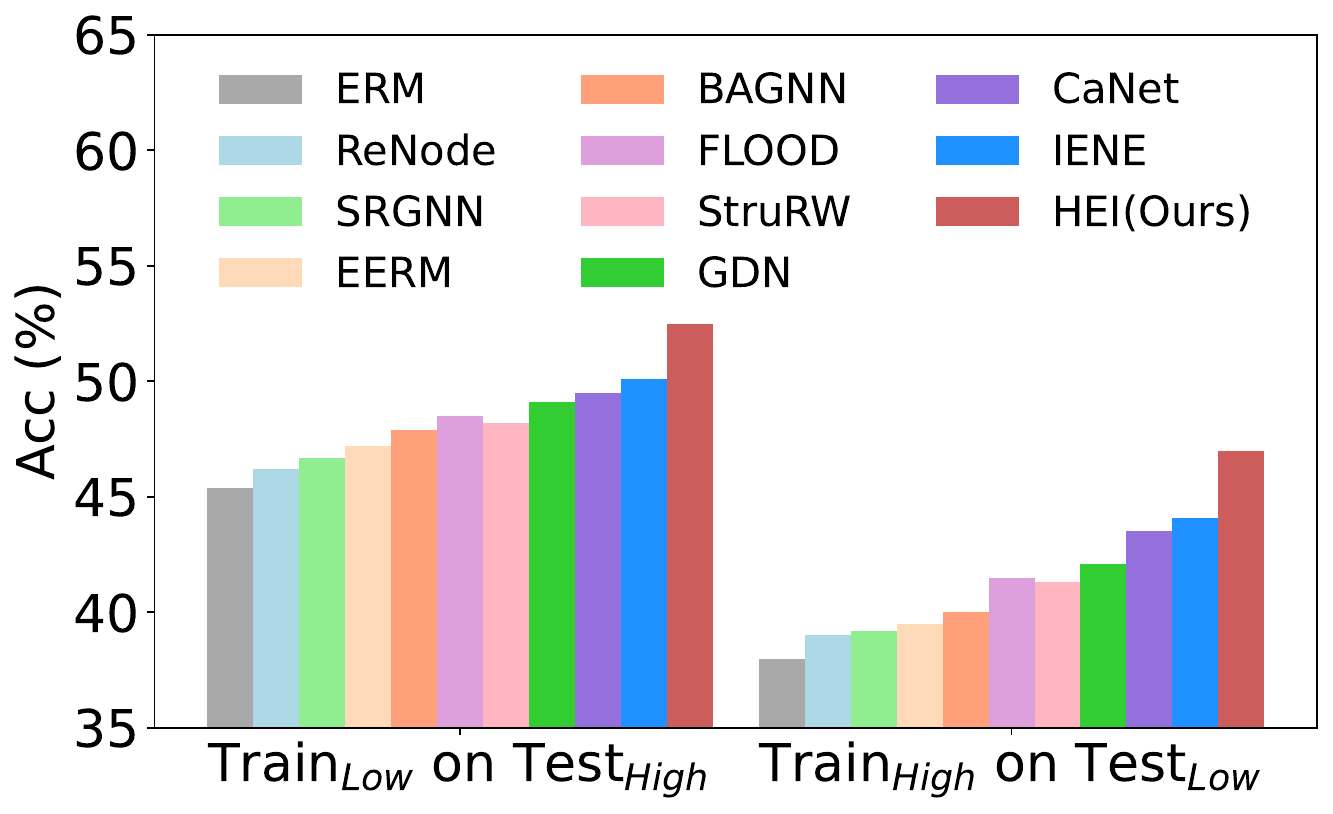}
}
\caption{Comparison experiments under Simulation Settings where exists severe distribution shift between train and test, including $Train_{High}$ on $Test_{Low}$ and $Train_{Low}$ on $Test_{High}$. We adopt the GloGNN++ as the backbone there.}
\Description{Comparison experiments under Simulation Settings where exist severe distribution shift between train and test, including $Train_{High}$ on $Test_{Low}$ and $Train_{Low}$ on $Test_{High}$. We adopt the GloGNN++ as the backbone there.}
\label{GloGNN_large}
\end{center}
\end{figure*}

\section{Implementation Details} \label{Implementation}
 For training details, we should warm up for some epochs to avoid the learned environments in the initial stage that are not effective, which may influence the optimization of models. After that, we adopt our proposed framework to learn an invariance penalty to improve model performance. For the range of parameters, we first execute experiments using basic backbones to get the best parameters of num-layers and hidden channels on different datasets. Then, we fix the num-layers and hidden channels to adjust other parameters, penalty weight$\lambda$) from $\left\{1e-, 1e-2, 1e-1, 1, 10, 100\right\}$, learning rate from $\left\{1e-2, 5e-3, 1e-3,5e-4,1e-4\right\}$ and weight decay from $\left\{1e-2,5e-3,1e-3\right\}$. We also provide parameter sensitivity of environment number $k$ in the paper. Moreover, the $\rho$ is a two-layer MLP with the hidden channel from $\left\{16,32,64\right\}$, and its learning rate should be lower than the backbone in our experiments, within the range from $\left\{ 5e-3, 1e-3, 5e-4,1e-4\right\}$.

\end{document}